\documentclass[11pt,oneside]{book}
\usepackage[margin=1.0in]{geometry}
\usepackage{setspace}
\usepackage[toc,page]{appendix}
\usepackage[none]{hyphenat} 
\usepackage{graphicx}
\usepackage{amsmath,amsfonts}
\usepackage{algorithm,algpseudocode}
\usepackage{tcolorbox}
\usepackage{caption}
\usepackage{textcomp}
\usepackage{hyperref}
\usepackage{array}
\usepackage{caption}
\usepackage{tikz}
\usepackage{graphicx}
\usepackage{subcaption}
\usepackage{sectsty}
\usepackage[toc, style=alttreegroup, nonumberlist,nopostdot]{glossaries}
\usepackage{notoccite}
\usepackage{tabularx}

\begin{document}
\frontmatter

\begin{titlepage}


\begin{center}
{\LARGE University of Sheffield}\\[1.5cm]
\linespread{1.2}\huge {\bfseries Dealing with Sparse Rewards in Reinforcement Learning }\\[1.5cm]
\linespread{1}
\includegraphics[width=5cm]{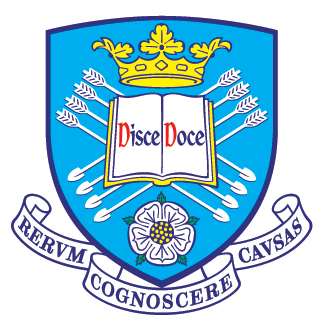}\\[1cm]
{\Large Joshua Hare}\\[1cm]
{\large \emph{Supervisor:} Eleni Vasilaki}\\[1cm]
\large A report submitted in fulfilment of the requirements\\ for the degree of MSc in Advanced Computer Science\\[0.3cm] 
\textit{in the}\\[0.3cm]
Department of Computer Science\\[2cm]
September 11, 2019 
\end{center}

\end{titlepage}


\newpage
\chapter*{\Large Declaration}

\setstretch{1.1} 

All sentences or passages quoted in this report from other people's work have been specifically acknowledged by clear cross-referencing to author, work and page(s). Any illustrations that are not the work of the author of this report have been used with the explicit permission of the originator and are specifically acknowledged. I understand that failure to do this amounts to plagiarism and will be considered grounds for failure in this project and the degree examination as a whole. 

\noindent Name: Joshua Hare\\[1mm]
\rule[1em]{25em}{0.5pt}

\noindent Signature: Joshua Hare\\[1mm]
\rule[1em]{25em}{0.5pt}

\noindent Date: 15/05/2019\\[1mm]
\rule[1em]{25em}{0.5pt}


\chapter*{\Large \center Abstract}
Successfully navigating a complex environment to obtain a desired outcome is a difficult task, that up to recently was believed to be capable only by humans. This perception has been broken down over time, especially with the introduction of deep reinforcement learning, which has greatly increased the difficulty of tasks that can be automated. However, for traditional reinforcement learning agents this requires an environment to be able to provide frequent extrinsic rewards, which are not known or accessible for many real-world environments. This project aims to explore and contrast existing reinforcement learning solutions that circumnavigate the difficulties of an environment that provide sparse rewards. Different reinforcement solutions will be implemented over a several video game environments with varying difficulty and varying frequency of rewards, as to properly investigate the applicability of these solutions. This project introduces a novel reinforcement learning solution by combining aspects of two existing state of the art sparse reward solutions, curiosity driven exploration and unsupervised auxiliary tasks. 

\tableofcontents
\newpage

\noindent \begin{minipage}[t]{1\linewidth}
\listoffigures
\end{minipage}
\begin{minipage}[b]{1\linewidth}
\listoftables
\end{minipage}

\glsaddall
\printglossary[title={Summary of Notation}]

\setstretch{1.1} 

\mainmatter
\chapter{Introduction}
Reinforcement learning is a multidisciplinary field combining aspects from psychology, neuroscience, mathematics and computer science, where an agent learns to interact with a environment by taking actions and receiving rewards. This is inspired from observations of chemical reward signals found in brains of humans and many other animal \cite{AnimalReward}, that dictate the behaviours of these animals. Video games have been a common benchmark for reinforcement learning (RL) throughout the existence of RL. From Samuel's checkers player \cite{Samuel} in 1959 to the game of Go \cite{AlphaGo}, Atari-2600 games \cite{ALE}, to most recently Deepmind's Starcraft II \cite{StarcraftII}, which has over $10^8$ possible available actions that the agent can take. The preference for Video games in RL, is easily understandable, as they provide an enclosed simulatory environment, often with easily accessible, quantifiable rewards that may not be available or difficult to measure in a real world environment. As it is common for many RL algorithms to take millions of training examples to effectively learn to interact with its environment, video games are useful in that they are computational simulations. This means that the environment is able to run faster than real-time, and many simulations can be performed simultaneously.\\

\noindent However, many modern approaches to RL such as Deep Q learning \cite{DQN} require, as stated previously, millions of examples in order to optimise due to some fundamental difficulties of the task. One of these is that often RL agents have no prior knowledge of the environment in this case video games, as often video games are modelled on and include real world objects such as ladders, and physics such as gravity. From \cite{HumanPrior} it was shown that humans perform significantly worse when prior knowledge is masked, and similarly, reinforcement agents perform better when given prior information about the environment.\\

\noindent Another fundamental difficulty of RL is that of a sparse reward signal provided by an environment. A Sparse reward signal is series of rewards produced by the agent interacting with the environment, where most of the rewards received are non-positive. This makes it extremely difficult for reinforcement learning algorithms to connect a long series of actions to a distant future reward. For extremely sparse rewards, the agent may never find a reward signal, and thus will never learn how to perform a given task. Human agents are not hindered by sparse reward signals and in fact are capable of achieving tasks over a entire lifetime with little to no reward through intrinsic motivation \cite{Curiosity}, traditional RL algorithm's learning is severely inhibited by sparse rewards.

This means that in order to automate a specific tasks with traditional reinforcement learning, the rewards received by the agent have to be carefully designed in order to maximise optimal behaviour. This is a problem as it requires that researchers and developers take time to fully understand the dynamics or optimal behaviour for an environment a priori. This takes time to implement, or worse the dynamics or optimal behaviour may unknown to the designers, so the rewards given to the agent may provide sub-optimal or non-functioning behaviour. By having an agent that can learn from sparse rewards, it allows designers to provide more abstract and long term goals, as well as being able tackle more complex tasks.\\

\noindent This project seeks to explore and contrast solutions to the inhibiting effect sparse rewards on RL agents, their implementations can be found on the following GitHub repository \\ \href{https://github.com/jhare96/reinforcement-learning/}{https://github.com/jhare96/reinforcement-learning/}.

\section{Aims and Objectives}

In this project the overall aims are:
\begin{itemize}
	\item Use reinforcement learning to successfully implement a working agent capable of playing video games
	\item Explore and contrast existing reinforcement methods to deal with sparse reward signals 
	\item Implement existing solutions for sparse reward reinforcement learning on a series of increasing difficulty tasks
\end{itemize}

\section{Overview of the Report}
This report begins with a literature survey starting with chapter \ref{chapter1}, which is a detailed introduction to reinforcement learning, specifically with the use of Markov Decision Processes. This chapter introduces some key terminology for RL and also introduces the mathematical framework used to evaluate and create RL algorithms as well the algorithms themselves. The literature survey continues on to introduce neural networks in chapter \ref{chapter2}, finally a final chapter \ref{chapter3} on deep neural networks use in RL, including state of the art neural network implementations related to solving RL with sparse rewards. 

Chapter \ref{chapter5} contains a brief description of the environments used to evaluate the agents implemented in this project.
Following this is chapter \ref{chapter6} dedicated to a detailed explanation on the implementations of the RL agents tested in this project, as well as preliminary analysis on their expected performance. The results of the experiments performed are presented in chapter \ref{chapter7} and finally the conclusions of the project are found in chapter \ref{chapter8}

\chapter{Reinforcement Learning Background } \label{chapter1}

\section{Markov Chains}
A Markov chain is a probabilistic model that connects are series of states $s$ in the set $\mathcal{S}$ via transition probabilities, i.e. moving from one state to another \cite{bishop2006pattern}. A state is some representation of a model, this could be a known behaviour or variable, or an unobserved or partially observed hidden representation of a system (Hidden Markov Model). The transition probabilities between the states in the Markov Chain hold the Markov property, that is the transition between the next state of a system $s_{t+1}$ is only dependent on the current state $s_t$ irrespective of the previous state $s_{t-1}$ \cite{markovchain}.
\begin{equation}
p(s_{t+1} | s_0,s_1,\hdots,s_t) = p(s_{t+1}|s_t)
\end{equation}

\section{Discrete Markov Decision Processes}
A Discrete Markov Decision Process (MDP) is a decision making process characterised by a discrete state and time Markov chain. In a MDP the set of states $\mathcal{S}$ are a representation of the environment e.g. game piece configuration on a board game \cite{MDP}. In an MDP at each time step a decision is made by taking an action $a_t$ from a set of actions available in state $s$, $a_t \in \mathcal{A}_s$. Taking the action $a_t$ produces a state transition modelled by the probability distribution $ \mathcal{P}^a_{ss'} = P(s_{t+1} = s' | s_t = s, a_t = a)$ \cite{Sutton2018}, following each state transition a real reward is received $r_t$ \footnote{The reward after transitioning from state $s_t$ to $s_{t+1}$ is sometimes denoted as $r_{t+1}$ \cite{Sutton2018}. The reasoning for the choice of $r_t$ is to be consistent with the modern deep reinforcement learning papers \cite{DQN},\cite{nature},\cite{RND},\cite{Curiosity_Scale}, which use this notation.} $\in \mathcal{R}$ characterised by the reward model $r_t = \mathcal{R}_s^a = (r_{t} | s_t = s, a_t = a)$.

\section{Reinforcement Learning in Markov Decision Processes}

Reinforcement learning is an optimisation process in which an agent learns a policy $\pi$ (a strategy in which determines what decisions will be made), by interacting with an environment such that it maximises the expected future reward.
For reinforcement learning in an MDP the optimisation process is to learn the optimal policy $\pi \approx \pi^*$ which maximises the cumulative expected future reward \cite{Sutton2018}, known as the return\footnote{Also referred to as $G_t$ in \cite{Sutton2018}} $R$. The return at time $t$, $R_t$ is the expected future reward from state $s_t$ onwards indefinitely for continuous tasks or until the final state $s_T$ for episodic tasks. 
\begin{equation}
R_t = \sum_{k=0}^{\infty}{r_{t+k}} \label{eq:R}
\end{equation}

\noindent A policy $\pi$ maps the state $s$ onto a probability distribution over the available actions in $\mathcal{A}_s$ \cite{Sutton2018}. The probability for each action $a \in \mathcal{A}_s$ is denoted by the term $\pi(a|s)$.
Value functions provide an estimation of the value of being in state $s$ or the value of taking an action $a$ in $s$ under a certain policy $\pi$. Commonly used in reinforcement learning are the value functions $V^\pi$ and $Q^\pi$, which denote the value and state-action value functions respectively.  

The value function for a given state $s$ under $\pi$ is given by the expected future reward of from starting at $s$ and following the policy $\pi$ until the terminal state. Often a discount factor $\gamma \in [0,1]$ is used dampen the reward signal from distant future actions so that the agent prioritises more immediate rewards (with the idea that immediate rewards are better than distant ones) and to deal with any uncertainty in the model \cite{SilverRL}.
Similarly the state-action value function is given by the expected future reward starting from $s$ taking action $a$ then following $\pi$ onwards. The value function and state-action value function are given by:

\begin{align}
V^\pi(s) &= \mathbb{E}_\pi[R_t | s_t = s] = \mathbb{E}_\pi\left[\sum_{k=0}^\infty{\gamma^k r_{t+k} } \,\middle\vert\, s_t = s\right] \label{eq:value-fn} \\
Q^\pi(s,a) &= \mathbb{E}_\pi[R_t | s_t = s, a_t = a] = \mathbb{E}_\pi \left[\sum_{k=0}^\infty{\gamma^k r_{t+k} } \,\middle\vert\, s_t = s, a_t = a \right] \label{eq:state-action-fn}
\end{align}

In order to solve reinforcement learning, the optimal policy $\pi^*$ is to be found, which is to say to find the strategy which yields the highest reward over all the states. A policy $\pi$ is considered better than another policy $\pi'$ if it's value function is greater over all states \cite{Sutton2018}, $V^\pi(s) > V^{\pi'}(s) \,\, \forall s \in S$. Therefore, this can be extended to define the optimal policy:
\begin{align}
\pi^* = \arg\max_{\pi} V^\pi(s) \,\, \forall s \in S \label{eq:policy1}
\end{align}

Since $V^*$ is the maximum expected reward from starting from state $s$ then it must be the maximum of the expected reward over all possible available actions \cite{Sutton2018}
$V^*(s) = \max_{a \in \mathcal{A}_s} Q^*(s,a)$. Hence, the optimal policy can be written in terms of the optimal state-action value function:
\begin{align}
\pi^* = \arg\max_{a} Q^*(s,a) \,\, \forall s \in S \label{eq:policy2}
\end{align}

\section{Dynamic Programming}
Dynamic programming is technique developed by Bellman to recursively solve a structural search problem (in this case a the optimal policy for a MDP), by storing the values of computed variables that reoccur during the recursive process \cite{dynamic-programming}. From equations \eqref{eq:value-fn} \& \eqref{eq:state-action-fn} it can be seen that the cumulative expected reward $R_t = \sum_{k=0}^{\infty}{r_{t+k}}$ is calculated at each time step $t$, which can be computationally demanding for large MDPs.\\
Using $\gamma^0 = 1$ \eqref{eq:value-fn} can be expanded \cite{EleniRL}:
\begin{flalign}
V^\pi(s) &= \mathbb{E}_\pi \left[ r_{t} + \gamma r_{t+1} + \gamma^2 r_{t+2} + \hdots \gamma^K r_{t+K} \,\middle\vert\, s_t = s \right] \,\, \text{Using Definition of $R_t$ \ref{eq:R}} \nonumber &&\\
&= \mathbb{E}_\pi\left[ r_{t} + \gamma R_{t+1} \,\middle\vert\, s_t = s\right]  \label{Expected_future_r} &&\\ 
&\text{conditioning on the next state $s'$ using law of total expectation} \nonumber &&\\
& = \sum_{s' \in \mathcal{S}} {P(s_{t+1} = s' | s_t = s)\mathbb{E}_\pi\left[ r_t + \gamma R_{t+1} \,\middle\vert\, s_t = s, s_{t+1} = s' \right]} \nonumber &&\\
&\text{then condition on the action $a$} \nonumber &&\\
&\sum_{a}{P(a|s)} \sum_{s' \in \mathcal{S}}{P(s' | s,a)\mathbb{E}_\pi\left[ r_t + \gamma R_{t+1}  \,\middle\vert\, s, s', a \right]} \nonumber &&\\
&\text{ then using addition theorem of expectations} \nonumber &&\\
&\sum_{a}{P(a|s)} \sum_{s' \in \mathcal{S}}{P(s' | s,a) \left[ \mathbb{E}_\pi\left[ r_t \,\middle\vert\, s, s', a \right] + \mathbb{E}_\pi\left[ R_{t+1} \,\middle\vert\, s, s', a \right] \right]} \nonumber &&\\
&\text{using the Markov property and definition of Reward model} \nonumber &&\\
&\sum_{a}{P(a|s)} \sum_{s' \in \mathcal{S}} {P(s' | s,a) \left[ \mathcal{R}^a_{s} + \mathbb{E}_\pi\left[ R_{t+1} \,\middle\vert\, s'\right] \right]} \nonumber &&
\end{flalign}

\noindent Thus the value function can be expressed recursively with the reward and transition model:
\begin{align}
&V^\pi(s) = \sum_{a}{\pi(a|s)} \sum_{s' \in \mathcal{S}}{\mathcal{P}^a_{ss'} \left[ \mathcal{R}^a_{s} + \gamma V^\pi(s') \right]} \label{eq:BellmanV}
\end{align}
\noindent A similar derivation can be done conditioning on the next state $s'$ and next action $a'$ or from \eqref{eq:BellmanV} using:
\begin{align}
&V^\pi(s) = \sum_{a \in A}{\pi(a|s)Q^\pi(s,a)} \,\, \implies Q^\pi(s,a) = \sum_{s' \in \mathcal{S}}{\mathcal{P}^a_{ss'} \left[ \mathcal{R}^a_{s} + \gamma V^\pi(s') \right]} \nonumber \\
&Q_{\pi}(s, a)=\sum_{s' \in S} \mathcal{P}_{ss'}^{a} \left[ \mathcal{R}_{s}^{a} + \gamma \sum_{a' \in \mathcal{A}} \pi(a' | s') Q_\pi(s', a') \right] \label{eq:BellmanQ}
\end{align}
These are known as the Bellman equations for stochastic processes. \\\\

\noindent Dynamic program requires a known MDP in order to estimate the optimal value function hence update the policy. The policy can be updated iteratively by computing the value function of a policy $\pi$ as according to the Bellman equations given above, then updating the policy using \eqref{eq:policy1} or \eqref{eq:policy2} depending on the value function used. 
\begin{figure}[H]
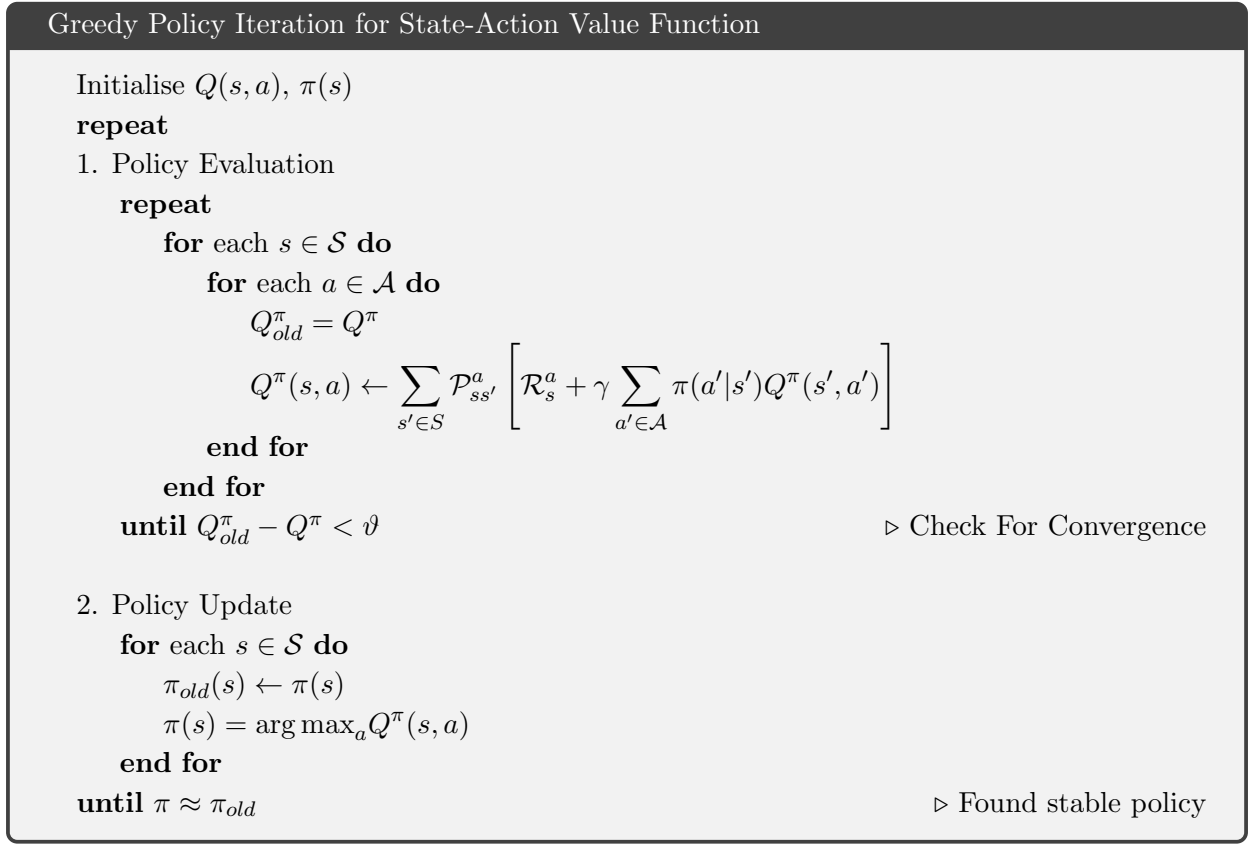

	\begin{tcolorbox}[title = Greedy Policy Iteration for State-Action Value Function]
		\begin{algorithmic}
			\State{Initialise $Q(s,a),\, \pi(s)$}
			\Repeat
			\\ \text{1. Policy Evaluation}
			\Repeat 
			\For {each  $s \in \mathcal{S}$}
			\For {each  $a \in \mathcal{A}$}
			\State $Q^\pi_{old} = Q^\pi$
			\State $\displaystyle Q^\pi(s,a) \leftarrow \sum_{s' \in S} \mathcal{P}_{ss'}^{a} \left[ \mathcal{R}_{s}^{a} +
			\gamma \sum_{a' \in \mathcal{A}} \pi(a' | s') Q^\pi(s', a') \right]$
			\EndFor
			\EndFor
			\Until{$Q^\pi_{old} - Q^\pi < \vartheta$} \Comment{Check For Convergence}
			\\
			\\ \text{2. Policy Update}
			\For {each  $s \in \mathcal{S}$} 
			\State $\pi_{old}(s)  \leftarrow \pi(s)$
			\State $\displaystyle \pi(s) = {\arg\max}_{a} Q^\pi(s,a)$
			\EndFor
			\Until{$\pi \approx \pi_{old}$} \Comment{Found stable policy}
		\end{algorithmic}
		
	\end{tcolorbox}
	\caption[Greedy policy iteration for state-action value function]{Greedy policy iteration for state-action value function algorithm, adapted from \cite{Sutton2018} } \label{fig:Greedy_Policy}
\end{figure}

\section{Model Free Reinforcement Learning}
Rather than learning the optimal policy for a known MDP structure, model-free reinforcement learning attempts to find the value function of being in state $s$ directly from experience. This is a crucial step for applying reinforcement learning to real world problems, as most environments have an unknown MDP or too large state-action space to computationally model. The cost of this approach is the agent has a less efficient learning process, i.e. many more samples required to optimally converge.

\subsection{Monte Carlo}
In Monte Carlo reinforcement learning is a model free method which learns the value function for episodic tasks. An episodic task is a sequence of sequential experiences $\left<s_t,a_t,r_t, \right>$ that \textbf{always} have a terminal state i.e. from $s_0,\hdots,s_T$. Most games can be defined as episodic tasks, an example being a game of chess always has a terminal state (a final board-piece configuration) after the game has finished through a win or draw. The reason Monte Carlo reinforcement learning can only learn from episodic tasks is that it approximates the expectation of future rewards in the value functions \eqref{eq:BellmanV} and \eqref{eq:BellmanQ} with an average of $R_t$ for all states visited, over each episode. The future reward $R_t$ can be determined online with dynamic programming however, the average state count can only be determined at the end of the episode. The average will tend to the true expectation as the number of samples approaches infinity due to the central limit theorem. Two common ways the average is calculated in Monte Carlo methods are the first visit evaluation and the every visit evaluation. In the first visit approach, the value function for a state $s$ is calculated only once per sample (episode) and is averaged over all samples. The every visit approach averages the value of the state for every visit including multiple visits per episode \cite{Sutton2018}. The Value function can be expressed via $V(s) = S(s) / N(s)$, where $S(s)$ is the sum of sample value estimates from state $s$ and $N(s)$ is the number of times the agent has visited state $s$. \\
The value function can be updated iteratively in Monte Carlo learning, as the mean of a series can be computed iteratively. For a dynamic problem it may be useful to calculate the running mean whilst forgetting older episodes, this is by introducing a dynamic variable $\alpha$ \cite{Sutton2018}:
\begin{equation}
V\left(s_{t}\right) \leftarrow V\left(S_{t}\right)+\alpha \left(R_{t}-V\left(S_{t}\right)\right) \label{MC}
\end{equation}
\noindent A disadvantage Monte Carlo method introduces is the inability to learn continuous tasks and the inability to update the value function before the end of an episode. This is not an issue for short time length episodes, but can greatly increase the time taken to find the optimal policy.

\subsection{Temporal Difference}
Temporal Difference combines the model free learning from the Monte Carlo method with the bootstrapping (expected future reward value estimation) from Dynamic Programming. This allows policy learning during episodes or continuous tasks, without the need to model the transition probabilities $\mathcal{P}^a_{ss'}$. In Temporal Difference learning, the optimal value function is found by iteratively updating the value function with the temporal error between the value predicted at $V(s_t)$ and the future reward $R_t = \sum_k \gamma^k r_{t+k}$ similar to Monte Carlo evaluation \eqref{MC}. This is done by sampling from the expected future reward from \eqref{Expected_future_r} and using current value function estimate instead of the true value function, as seen in Dynamic Programming \cite{Sutton2018}. The temporal error is then:
\begin{equation}
\delta_t = r_t + \gamma V(s_{t+1}) - Vs_t)
\end{equation}

$V(s)$ can be iteratively updated using the temporal error $\delta_t$, this is known as $TD(0)$:

\begin{equation}
V(s) \leftarrow V(s) + \alpha \left[r_t + \gamma V(s_{t+1}) - V(s_t) \right] \label{TDV}
\end{equation}

Using the state-action value function in \eqref{TDV} results in the on-policy method commonly known as SARSA (state, action, reward, state, action)
\begin{equation}
Q(s,a) \leftarrow Q(s,a) + \alpha \left[r_t +  \gamma  Q(s_{t+1},a_{t+1}) - Q(s_t, a_t) \right] \label{SARSA}
\end{equation}
A bridge between Temporal Difference and Monte Carlo methods can be extended first with an $n$-step Temporal Difference error TD($n$) which is defined as the temporal error between state $V(s_t)$ and $V(s_{t+n})$ defined using the $n$-step future return $R_{t}^{(n)}$ \cite{Sutton2018}:
\begin{flalign}
&R_{t}^{(n)}=r_{t}+\gamma r_{t+1}+\ldots+\gamma^{n-1} r_{t+n-1}+\gamma^{n} V\left(s_{t+n}\right) \label{n-step} \\
&\delta_t^{(n)} = R_{t}^{(n)} - V(s_t)
\end{flalign}
The optimal number of steps $n$ depends on the length of the task and the number of states, finding a general solution to this problem is done by further extending the bridge to Monte Carlo is the $TD(\lambda)$ approach. This is where all the expected rewards \eqref{n-step} are computed from $n=0,\hdots T$ and combined geometrically via \cite{Silverlambda}:
\begin{equation}
R_{t}^{\lambda}=(1-\lambda) \sum_{n=1}^{\infty} \lambda^{n-1} R_{t}^{(n)} \label{eq:lambda}
\end{equation}

The $\lambda$ approach can also be applied to SARSA, SARSA($\lambda$). \\

\noindent   Q Learning(an off-policy version of SARSA) was developed by Watkins \cite{Watkins1992}. Here instead of approximating the true state-action value function for a given policy $Q^\pi(s,a)$ with the current estimate, the optimal state-action value function is approximated $Q^*(s,a)$.
\begin{equation}
Q(s,a) \leftarrow Q(s,a) + \alpha \left[  r_{t} + \gamma \max_a Q(s_{t+1},a) - Q(s_t, a_t) \right] \label{QLearning}
\end{equation}
Q Learning suffers from a maximisation bias, that is that it overestimates the values in the state-action function $Q(s,a)$ i.e. with a positive bias. A solution was proposed in \cite{DoubleQ-learning} to use two state-action value functions, one to predict the next action to be taken and another to predict the value estimation of the action. \eqref{QLearning} then becomes:
\begin{equation}  \label{Double-QLearning}
\begin{aligned}
Q_A(s,a) \leftarrow Q_A(s,a) + \alpha \left[  r_t + \gamma Q_B(s',{\arg\max}_a Q_A(s',a)) - Q_A(s, a) \right] \\
Q_B(s,a) \leftarrow Q_B(s,a) + \alpha \left[  r_t + \gamma Q_A(s',{\arg\max}_a Q_B(s',a)) - Q_B(s, a) \right] 
\end{aligned}
\end{equation}
Both $Q_A(s,a)$ and $Q_B(s,a)$ are used as the to select the action at time $t$ and at each optimisation step either A or B are updated via random selection. \\

\noindent Q Learning also suffers from poor exploration as the approximation to optimal value is being used as the policy hence often limiting the exploration to states of known rewards. A solution to this is to use a decaying $\varepsilon$ - greedy policy to enable initial exploration, then use optimal policy once the states have been sufficiently explored.
\begin{equation}
\varepsilon - greedy \left\{\begin{array}{ll} { a_t = \text {random}\, a \in \mathcal{A}} & {\text{with probability}\,\, \varepsilon } \\
{a_{t}=\max _{a} Q^{*}\left(s, a \right)} &  \text{with probability}\,\,{1 - \varepsilon }\end{array}\right.
\end{equation}

\subsection{Policy Gradient}
Rather than learning the optimal policy through the maximisation over value function approximations \ref{eq:policy1} or \ref{eq:policy2} (which can be expensive for high dimensional state-actions spaces), the policy can be learnt directly under the condition that the policy is parametrised by a differentiable function. The policy is learnt by calculating the gradient of the current policy estimate $\nabla \pi(a|s,\theta)$ \cite{Sutton2018}. This allows us to learn, stochastic policies, continuous action policies and should theoretically provide more stable convergence properties \cite{SilverRL} over value-based methods such as Q-Learning. However learning the policy directly can reduce the sample efficiency of the agent taking longer to converge to the optimal policy. \\
The policy gradient theorem shows that maximising the gradient of the value function is equivalent to maximising the gradient of the policy multiplied by the state-action value function all under the expectation of the policy \cite{Sutton2018}. This gives the gradient objective function $J(\theta)$ (with the objective of maximising the expected future reward) as

\begin{equation}
\nabla_{\theta} J(\theta)=\mathbb{E}_{\pi_{\theta}} \left[ Q^{\pi}(s,a) \nabla_{\theta} \log \pi(a|s, \theta) \right] \label{eq:policy_grad}
\end{equation}
Using the definition of $Q^\pi(s,a)$ \ref{eq:state-action-fn}
\begin{equation}
\nabla_{\theta} J(\theta)=\mathbb{E}_{\pi_{\theta}} \left[R_t \nabla_{\theta} \log \pi(a|s, \theta) \right] \label{eq:policy_grad_G}
\end{equation}
As $Q^{\pi}(s,a)$ is not dependent on $\theta$ a baseline metric $b(s_t)$ can be subtracted from equations \ref{eq:policy_grad} \& \ref{eq:policy_grad_G} as long as it is constant w.r.t. $a$.
\begin{equation}
\nabla_{\theta} J(\theta)=\mathbb{E}_{\pi_{\theta}} \left[(R_t - b(s)) \nabla_{\theta} \log \pi(a|s, \theta) \right] 
\end{equation}
\subsubsection{Actor Critic}
The Actor-Critic family of policy gradient methods involves combining the baseline policy gradient method with the bootstrapping method from temporal difference. For the Advantage Actor-Critic the baseline function used is the Advantage function 
\begin{equation}
A(s_t,a_t) = Q(s_t,a_t) - V(s_t) = r_t + \gamma V(s_{t+1}) - V(s_t). \label{eq:ADV}
\end{equation} Expressing the Advantage function in terms of only $V^\pi$ allows for the use of a single set of parameters to define the advantage function and reducing the dimensionality of the function being learnt. This has the benefits of being easier to learn as well as decreasing computational complexity, especially for large $|\mathcal{A}|$ \cite{GAE}. 
\begin{equation}
\begin{aligned}
\nabla_{\theta} J(\theta)&=\mathbb{E}_{\pi_{\theta}} \left[(r_t + \gamma V(s_{t+1}) - V(s_t)) \nabla_{\theta} \log \pi(a_t|s_t, \theta) \right] \qquad \text{where $a_t \sim \pi$} \\
&=\mathbb{E}_{\pi_{\theta}} \left[\delta_t \nabla_{\theta} \log \pi(a_t|s_t, \theta) \right]
\end{aligned}
\end{equation}
By using $V(s)$ as a baseline as well as to predict the TD error, value function provides the estimate of the quality of the action taken by the policy. Hence, it is said to be acting as a critic to the policy (the actor), thus giving the name Actor-Critic. 

\subsubsection{Generalised Advantage Estimation}
Just as the TD $n$-step returns estimator can be combined in TD($\lambda$), the advantage function estimation can be calculated similarly. Following \ref{eq:ADV}, the $n$-step advantage estimator can be defined as 
\begin{equation}
A^{n}_t = \sum_{l=0}^{n-1} \gamma^l \delta_{t+l} = \sum_{l=0}^{n-1} \gamma^l r_{t+l} + \gamma^{n} V(s_{t+n}) - V(s_t)
\end{equation}

\begin{align}
\intertext{Following the derivation from \cite{GAE}, $A^\lambda_t$ can be derived by first exponentially averaging each $n$-step advantage estimate}
A^{\lambda}_t &= (1-\lambda)\left(A_{t}^{(1)}+\lambda A_{t}^{(2)}+\lambda^{2} A_{t}^{(3)}+\ldots\right) \nonumber \\
A^{\lambda}_t &= (1-\lambda)\left(\delta_t + \lambda(\delta_t + \gamma\delta_{t+1}) +\lambda^2(\delta_t + \gamma\delta_{t+1}+ \gamma^2\delta_{t+2}) + \hdots\right) \nonumber \\
\intertext{Factoring by $\delta_t$}
A^{\lambda}_t &= (1-\lambda)\left(\delta_t(\lambda + \lambda^2 + \hdots) + \gamma\delta_{t+1}(\lambda^2 + \lambda^3 + \hdots) + \hdots \right) \nonumber \\
\intertext{Using the geometric series solution \quad $\displaystyle \sum_{k=0}^{n-1} a r^{k}=a\left(\frac{1-r^{n}}{1-r}\right)$}
A^{\lambda}_t &= (1-\lambda)\left(\delta_t\left(\frac{1}{1-\lambda}\right) + \gamma\delta_{t+1}\left(\frac{\lambda}{1-\lambda}\right) + \gamma^2\delta_{t+2}\left(\frac{\lambda^2}{1-\lambda}\right) + \hdots \right) \nonumber
\end{align}
then 
\begin{equation}
A^{\lambda}_t = \sum_{l=0}^{\infty}(\lambda\gamma)^l \delta_{t+l}  \label{eq:GAE}
\end{equation}

\chapter{Neural Networks} \label{chapter2}
Artificial neural networks are neurologically inspired computational frameworks, which simulate neuron activity as a weighted summation over a series of connected input neurons. There are many different neural networks topologies that are commonly used, but most consist of a series of connected groups of neurons known as layers, which are fed stimulatory input through an input layer. An output layer (most often the final layer of the network) provides an estimation of some target value that is attempted to being modelled by the network. Neural networks were first developed the in the 1940s \cite{Hebb}, however have seen a large increase in usage in recent years, due to increased computational power of modern computers. This allows larger, hence more powerful networks to be trained within a reasonable time frame. 
\section{Multi-Layer-Perceptron}
Inspired by the perceptron algorithm Hinton et al created a feed-forward multi-layer perceptron (MLP) \cite{MLP}. MLPs consist of distinct layers of nodes, where each node's value is a summation of the previous layer $l-1$ with the synaptic weights of the current layer $w^{l}_{ij}$ under a non-linear activation function $g$. MLPs can approximate any function with an infinitely sized hidden layer \cite{HORNIK} and can be trained to minimise a given loss function via the backpropagation algorithm \cite{lecun-88}. Non-linear activation functions are a key process of the MLP's ability to approximate any function and they are required to be differentiable in order to update the weights via backpropagation. The value of a node of a hidden layer $l$ is given by:
\begin{equation}
h^{l}_i = g \left(\sum_j w^l_{ij} h^{l-1}_j \right) = g \left(W^l h^{l-1} \right) \qquad \text{where $W \in \mathbb{R}^{n\times m}$} \label{eq:MLP}
\end{equation}
A loss function is defined by the final output layer $\hat{y}$ and the target value $y$ given by the model (weights) $\theta$, $\mathcal{L}(\hat{y},y|\theta)$. As the output layer is a composite function the weights can be updated by differentiating the weights w.r.t the loss function using the chain rule.
\begin{equation}
\frac{\partial \mathcal{L}}{\partial w^l_{ij}} = \frac{\partial \mathcal{L}}{\partial z^{l}_i} \frac{\partial z^{l}_{i}}{\partial w^{l}_{ij}}
= \frac{\partial \mathcal{L}}{\partial h^{l}_{i}} \frac{\partial h^{l}_{i}}{\partial z^{l}_{i}} \frac{\partial z^{l}_{i}}{\partial w^{l}_{ij}} 
\end{equation}
where $z = \sum_j w^l_{ij} h^{l-1}_j $. As the loss w.r.t the $h^l$ is dependent on the next layer $\frac{\partial \mathcal{L}}{\partial h^{l}_{i}}$ can be expressed recursively:
\begin{align}
\frac{\partial \mathcal{L}}{\partial w^l_{ij}}  &= 
\left(\sum_j \frac{\partial \mathcal{L}}{\partial z^{l+1}_{j}} \frac{\partial z^{l+1}_{j}}{\partial h^l_i} \right) \frac{\partial h^{l}_{i}}{\partial z^{l}_{j}} \frac{\partial z^{l}_{i}}{\partial w^{l}_{ij}}
&&\text{Since} \,\, \frac{\partial z^{l}_{i}}{\partial w^{l}_{ij}}  = h^{l-1}_i  \text{and} \,\, \frac{\partial h^{l}_{i}}{\partial z^{l}_{j}} = \frac{\partial g(z^{l}_{i})}{\partial z^{l}_{j}} \nonumber \\ \nonumber \\
&= \left(\sum_j \frac{\partial \mathcal{L}}{\partial z^{l+1}_{j}} \frac{\partial z^{l+1}_{j}}{\partial h^l_i} \right) \frac{\partial g(z^{l}_{i})}{\partial z^{l}_{j}} h^{l-1}_i 
\end{align}

\section{Convolutional Neural Networks}
Convolutional neural networks (CNN) are a visual cortex inspired neural network similar to the MLP as described above. They differ by the replacing the linear summation by a convolutional operation across the input. The convolution operation can be physically interpreted as an area overlap between two functions \cite{convolution}, in the context of neural networks it can be loosely interpreted as comparing a signal (image in video games) to a set of learnt filters which have a feature representation of different components of a signal. Using a convolutional layer allows the network to learn spatially or temporally invariant features across an input signal as the same filter is used across different sections of the input. In the context of images, this is equivalent to learning specific features, for example the features of a car regardless of the location of the car within the input. The convolution operation is defined by,
\begin{equation}
\begin{aligned}
&f(t) * g(t) = \int_{-\infty}^{\infty} f(\tau)g(t-\tau) d\tau &&\text{For continuous functions} \\
&x[m] * y[m]=\sum_{n=-\infty}^{\infty} x[n] y[m-n]  &&\text{For discrete functions} \\
\end{aligned}
\end{equation}
The hidden layer component $h^l_{ij}$ for a 3D fully convolutional layer with a single filter is given by, 
\begin{align}
&h^l_{ij} = g\left(w^l_{ijk} * h^{l-1}_{ij}\right)= g\left(\sum_x \sum_y \sum_z \, w^{l}_{x,y,z} h^{l-1}_{i-x,j-y,k-z}\right) &&\text{for a singular $k\times k\times c$ filter $w^l_{ijk}$}. \label{eq:CNN}
\end{align}

\section{Recurrent Neural Networks}
CNN and MLP are both examples of feed-forward neural network architectures. Specifically, these are both discrete acyclic graphs, meaning that each node in the graph is not connected to itself via any path. This reduces the capability of these networks for modelling sequential data, when the features $x \in \mathbb{R}^{n}$ are fed into the network at each time step $t$. This is because, in acyclic graphs there in no modelled relation between each pass of the network. Recurrent Neural Networks (RNN), solve this problem by combining the hidden state $h_t$ of the feature $x_t$, with the hidden state $h_{t-1}$ of the previous feature $x_{t-1}$. The hidden layer at time $t$ for a single-hidden layer RNN is given by the equation,
\begin{equation}
h_t = g\left(W_x \, x_t + W_h \, h_{t-1}  \right) \qquad \text{where $ \,\, W_x \in \mathbb{R}^{m \times n}$, $W_h \in \mathbb{R}^{m \times m}$}. \label{eq:RNN} 
\end{equation}

\noindent The output $y_t$ is defined by an additional MLP layer, and can be produced every time-step in a many-to-many configuration or at the final time-step $T$ in a many-to-one configuration. Another configuration includes that of one-to-many where the output $y_{t-1}$ is fed into \ref{eq:RNN} instead of the previous hidden state $h_{t-1}$. 

As the hidden state $h_t$ hence output $y_t$ is conditioned of the previous step, the backpropagation algorithms flows backwards through the time steps $T \rightarrow t_0$, hence is known as backpropagation through time. However the drawbacks of this simple recurrent network is that of the vanishing gradients problem, where the gradients decrease in size from $T$ to $t_0$, hence simple RNNs can struggle to learn from long sequential data. This problem is reduced by the introduction of Long Term Short Memory (LSTM) recurrent networks which use multiple gates, in order to learn to extract more useful long-term important features. 

\section{Activation Functions}
Many different non-linear activation functions are used for artificial neural networks and bring different strengths and weakness depending on the usecase. A brief description of each activation along with the relevant formula is given below. 

The hyperbolic and logistic activation function families have been a popular choice throughout the history of neural networks. The two most common choices however are the tanh and sigmoid functions, these are given by the equations, 
\begin{align}
&g(x) = tanh(x) = \frac{\left(e^{x}-e^{-x}\right)}{\left(e^{x}+e^{-x}\right)} &&\text{tanh} \label{eq:tanh} \\ \nonumber \\
&g(x) = \sigma(x) =\frac{1}{1+e^{-x}} &&\text{sigmoid}. \label{eq:sigmoid}
\end{align}
The tanh and sigmoid function constrain the value of $x$ to within the boundaries $\left[0,1\right]$, $\left[-1,1\right]$ respectively. This can lead to vanishing gradients \cite{Data_Mining} for deeper neural networks when training with backpropagation. This is because the gradient is restricted to between the values of $\left[0,1\right]$, since backpropagation uses the chain rule, these gradients can quickly tend towards zero resulting in very small weight updates, as well as  numerical error effects dominating the weight updates. \\

\noindent Rectified linear unit (ReLU) is perhaps the most popular activation function due to its simplicity, speed and often performance increase. It is given by the equation, 
\begin{equation}
g(x)=\left\{\begin{array}{ll}{0} & {\text { for } x \leq 0} \\ {x} & {\text { for } x>0}\end{array}\right. \qquad \text{ReLU.} \label{eq:ReLU}
\end{equation}
The performance increase has often been attributed to the increase in sparsity of the connections that carry any non-zero value, this has been observed to be between 50\%-80\% in \cite{sparse-relu} for deep neural networks. Sparsity of neural connectivity is seen as a more biologically plausible model, as it is estimated that neurons in the human brain have a sparsity of between 90\%-99\% \cite{sparse-relu}. This intuitively makes sense as for random weight initialisation with mean zero should produce 50\% sparsity, as the sets of positive and negative real numbers are equal in size. The vanishing gradient problem is negated as the gradient of the ReLU function is either 1 for positive $x$ or zero otherwise
\begin{equation*}
\frac{\partial g(x)}{\partial x} =\left\{\begin{array}{ll}{0} & {\text { for } x \leq 0} \\ {1} & {\text { for } x>0}\end{array}\right.
\end{equation*}, hence the gradients do not reduce in size over multiple layers. However the downside is that ReLU units can enter a value such that the neuron is never activated again, this is known as the dying ReLU problem and is solved by replacing the zero value in \ref{eq:ReLU} by some function of $x$. For example it is replaced by $\alpha x$ in the leaky ReLU and $a(e^x -1)$ in the Exponential Linear Unit (ELU) functions. 
\begin{align}
&g(x)=\left\{\begin{array}{ll}{\alpha x} & {\text { for } x \leq 0} \\ {x} & {\text { for } x>0}\end{array}\right.  &\text{leaky ReLU.} \label{eq:leakyReLU} \\
&g(x)=\left\{\begin{array}{ll}{a(e^x -1)} & {\text { for } x \leq 0} \\ {x} & {\text { for } x>0}\end{array}\right.  &\text{ELU.} \label{eq:ELU}
\end{align} \\

\noindent The softmax activation function produces a normalised probability distribution over each feature in an input vector $x \in \mathbb{R}^d \rightarrow  z \in \mathbb{R}^d$, such that the probability of $z_i$ is is proportional to the exponent of the input value $x_i$ \cite{bishop2006pattern}. This is explicitly calculated by the formula, 
\begin{align}
&g(x_i) = \frac{e^{x_i}}{\sum\limits_{k \in d}^{}e^{x_k}} &&\text{softmax}. 
\end{align}
As the nature of this activation function this is generally used on the final layer of the network for multinomial logistic regression.

\section{Loss Functions}
In order to calculate the gradients a distance metric between the desired output $y \in \mathbb{R}^N$ and the actual output $\hat{y} \in \mathbb{R}^N$ must be defined. Mean squared error is a common distance metric which penalises the difference between two vectors by the mean of the squared distance between each feature 
\begin{equation}
MSE = ||y-\hat{y}||^2 = \frac{1}{N} \sum_{i=0}^N (y_i-\hat{y_i})^2.
\end{equation}
A better comparison between two probability distributions than the mean squared error of the distributions is the cross entropy. Entropy is a measurement of uncertainty in a probability distribution. Entropy is highest when a distribution is uniform and lowest as the distribution approaches the Dirac delta function. For example the uncertainty is high for the outcome of a fair sided die roll, hence it has high entropy. For a discrete probability distribution entropy is given by,
\begin{equation}
H(x) = -\sum_i p(x_i) \log{p(x_i)}.
\end{equation}
Information theory states that a message contains high information when the outcome is unpredictable \cite{shannon-entropy}, as you have gained information about the outcome. Information is lowest when the outcome is known and predictable, as you already had the information about what the outcome would be. Entropy then can be seen as the amount of information in a signal. The Kullback-Leibler (KL) divergence can be seen as the amount of information needed to encode the information in the probability distribution $p$ with another distribution $q$ \cite{cross-entropy}, it is lowest when $q=p$ and hence can be seen as an error between these two probability distributions. Formally this is the expected log difference between the two distributions, 
\begin{equation}
D_{KL}(p||q) = \mathbb{E}_x\left[\log{p} - \log{q}\right] =  \sum_i p(x_i)\log{p(x_i)} - p(x_i)\log{q(x_i)} = \sum_i p(x_i)\log\left(\frac{p(x_i)}{q(x_i)}\right). \label{eq:KL}
\end{equation}
However this is not a true distance metric because it is not symmetrical $D_{KL}(p||q) \neq D_{KL}(q||p)$, nor does it satisfy the triangle equality \cite{cross-entropy-triangle}.

The second term in equation \ref{eq:KL}, $-\sum_i p(x_i)\log{q(x_i)}$ is also known as the cross entropy between $p$ and $q$. For a fixed $p$, for example the labels for a classification task, minimising the cross entropy is equivalent to minimising the KL-divergence. 
Minimising both the mean squared error and cross entropy are equivalent to maximising the likelihood of the data given the parameters \cite{bishop2006pattern}. 

\section{Optimisation Methods}

As backpropagation is a gradient based optimisation process, that minimises the parameters in some loss space $L$. Using gradient based optimisation allows the parameters to approach at least a local minimum for convex problems as well as stationary points for non-convex problems \cite{non-convex}. As neural networks require very large amount of data points compared to other Machine Learning algorithms, updating the gradient based on the average over all data points in a data set is impracticable. Thus, Stochastic Gradient Descent (SGD) methods are often used to increase training speed. \\

\noindent Using SGD methods is especially important as all data points cannot be accessed at any time, as it is being continually generated by the agent. SGD, however requires heavy tuning in order to find a optimal learning rate $\eta$ for different tasks, such that the parameters do not get stuck in local minima and preferable one that converges with as few updates as needed. Because of this modified SGD algorithms are often used throughout machine learning, especially with neural networks. 

\begin{figure}[H]
	\centering
	\includegraphics[width=.6\linewidth]{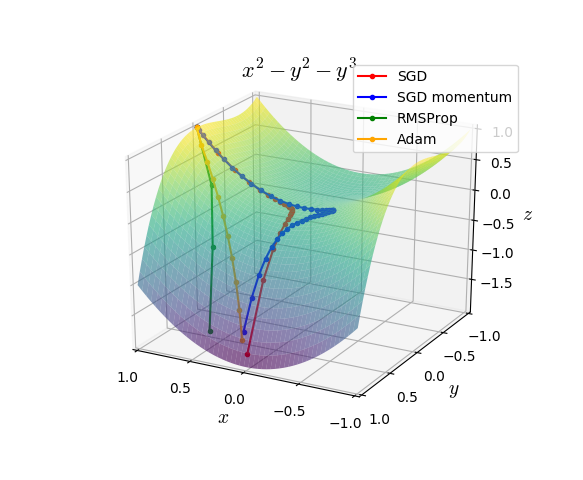}
	\caption[Visual representation of stochastic gradient optimisers]{Visual representation of stochastic gradient optimisers} \label{fig:opt}
\end{figure}

\subsection{Momentum based SGD}
In order to reduce stochasticity of the SGD updates, the concept of momentum of gradient updates was first used in \cite{Momentum}. Here the analogy is that of a particle rolling down a hill, it experiences a force pulling on it which increases its momentum. When the gradients are steeper the particle accelerates, and as the hill flattens it decelerates. This stabilises the direction of the gradient updates as the momentum makes it less susceptible to large deviations caused by a single minibatch update. If the optimiser has a large enough momentum it can cause it to escape suboptimal local minima. However, too large momentum can cause the optimiser to overshoot minima thus being slower to optimally converge, as seen in Figure \ref{fig:opt}.

\subsection{RMSProp}
Root Mean Squared Propagation (RMSProp) \cite{RMSProp}, adapts the magnitude of the update for each parameter by dividing by a running average of the root mean squared gradient $v=\bar{W}^2$, this allows different features to be learnt at different scales, thus reducing the need to finely tune the learning rate. For a neural network layer weighting $W$, the update rule can be written as

\begin{equation}
\begin{aligned}
&v_t \leftarrow \beta v_{t-1} + \left(1 - \beta\right) \left(\frac{\partial J}{\partial W_t}\right)^2 \\
&W_{t+1} \leftarrow W_t - \frac{\eta}{\sqrt{\vphantom{\beta^K} v_t}} \left(\frac{\partial J}{\partial W_t}\right)^2.
\end{aligned} \label{eq:RMSProp}
\end{equation}

\noindent Where $\beta$ is the decay rate of the average. 

\subsection{Adam}
Adaptive Moment Estimation (Adam) \cite{Adam} combines the adaptive magnitude updates from RMSProp \ref{eq:RMSProp} with momentum based gradient descent \cite{Adam}, with bias correction for momentum and adaptive moment (mean square of the gradient) 

\begin{equation}
\begin{aligned}
&m_t \leftarrow \beta_m m_{t-1} + \left(1 - \beta_m\right) \left(\frac{\partial J}{\partial W_t}\right) \qquad\text{momentum} \\
&v_t \leftarrow \beta_v v_{t-1} + \left(1 - \beta_v\right) \left(\frac{\partial J}{\partial W_t}\right)^2 \qquad\text{adaptive moment}\\
&\widehat{m}_{t} \leftarrow m_{t} /\left(1-\beta_{m}^{t}\right) \qquad\text{bias correction for momentum} \\
&\hat{v}_{t} \leftarrow v_{t} /\left(1-\beta_{v}^{t}\right)  \qquad\text{bias correction for adaptive moment} \\
&W_{t+1} \leftarrow W_{t}-\eta \, \frac{\widehat{m}_{t}}{\sqrt{\vphantom{\beta^K} \widehat{v}_{t}}+\epsilon} 
\end{aligned} \label{eq:Adam}
\end{equation}
Where $\beta_m^t, \,\beta_v^t$ are the decay rates with exponent of the number of updates $t$, and $\epsilon$ is a small scalar value to prevent division by zero \cite{Adam}.

\chapter{Deep Reinforcement Learning} \label{chapter3}
In video games it is common for the set of states $\mathcal{S}$ to be given as a pixel array of the game screen $s \in \mathbb{R}^{h\times w\times c}$, for a single colour frame of size $84\times84$ there are $84^2\times(255)^3 \approx 1\times10^{11} $ different image combinations. Most video games will likely only use a tiny fraction of this image space, but it is still impractical to model the value functions for each state separately. A solution is to use a differentiable function to model $V(s)$, this is known as value function approximation \cite{Sutton2018}. Deep Reinforcement Learning refers to using a deep neural network as a non-linear value function or policy approximator.

\section{Traditional Deep Reinforcement Learning}

\subsection{Deep Q Learning}
The Deep Q learning algorithm proposed by \cite{DQN} is essentially the Q learning algorithm described previously \eqref{QLearning}, with two additional key features that stabilise the policy iterations for the use of neural networks. The two additional features are experience replay, and fixed target policy iteration. If using a gradient based iterative optimisation process as done via back-propagation in neural networks, it is important that the data is independent and identically distributed (i.i.d.). This is done as to avoid sampling bias from correlated inputs, which can cause the gradient to get stuck in a non-optimal local maxima (as we are performing gradient ascent to maximise the expected future reward). Experience replay is a method to help decorrelate the sequential experiences gained from dynamic programming and model free reinforcement learning methods. This is done by storing experiences, a tuple of $\left<s_t,a_t,r_t,s_{t+1}\right>$ into a list of experiences known as the replay memory. Batch samples can be drawn randomly from the replay memory which provide $\sim $ i.i.d. for a large replay length. Deep Q Learning also solves the partially observable MDP problem by providing the input of the neural network with 4 concatenated previous states (greyscaled images for CNN).\\
The Deep Q Learning algorithm is expressed below : \\
\begin{figure}[h!]
	\begin{tcolorbox}[title = Deep Q Learning Algorithm with Experience Replay and $\epsilon$ - greedy Policy for Atari Environments]
		\begin{algorithmic}
			\State{Initialize replay memory $\mathcal{D}$ to capacity $N$}
			\State{Initialize action-value function $Q$ with random weights}
			\State{k - number of frames to stack}
			\For {episode $1\hdots M$}
			\State{$s_{1}=\left\{x_{1},x_{1},...\right\} \qquad \text{stack initial frame k-times}$}
			\For {time $t\hdots T$}
			\State{With probability $\varepsilon$ select a random action $a_t$}
			\State{otherwise $a_{t}=\max _{a} Q^{*}\left(s_{t}, a | \theta\right)$}
			\State{Execute action $a_t$ in emulator and observe reward $r_t$ and image $x_{t+1}$}
			\State{$s_{t+1}= \left\{x_{t-k+1},...,x_{t},x_{t+1}\right\}\, \text {and add experience to }\,\,\mathcal{D} \leftarrow \left<s_{t}, a_{t}, r_t,  s_{t+1}\right>$ }
			\State{Sample random minibatch of transitions $\left<s_{j}, a_{j}, r_{j}, s_{j+1}\right>$ from $\mathcal{D}$}
			\State{Set $y_{j}=\left\{\begin{array}{ll}{r_{j}} & {\text { for terminal }
					s_{j+1}} \\ {r_{j}+\gamma \max _{a^{\prime}} Q\left(s_{j+1}, a^{\prime} | \theta\right)} & {\text { for non-terminal } s_{j+1}}\end{array}\right.$} \\
			\State{$\nabla_{\theta_{i}} L_{i}\left(\theta_{i}\right)=\mathbb{E}_{\pi}\left[\left(r+\gamma \max _{a'} Q(s', a'| \theta_{i-1})-Q\left(s, a | \theta_{i}\right)\right) \nabla_{\theta_{i}} Q\left(s, a | \theta_{i}\right) \,\middle\vert\, s, a, s'\right]$}
			\State{Every $C$ steps $\theta_{i-1} = \theta_i$}
			
			\EndFor
			\EndFor
		\end{algorithmic}
		
	\end{tcolorbox}
	\caption[Deep Q Learning algorithm]{Deep Q Learning algorithm, adapted from \cite{DQN} } \label{fig:DQN}	
\end{figure}

\subsection{Advantage Actor Critic}
Asynchronous Advantage Actor-Critic (A3C) \cite{Async} is an algorithm where multiple actors interact and learn from their own environment asynchronously. Each actor has it's own local policy parameters which it uses to calculate the gradients of the policy and value function over n-steps, this is done using TD($n$). The gradients calculated by each actor are asynchronously applied to shared global parameters, and after each $n$-step rollout the local parameters are updated from the global parameters. Running multiple environments allows the agent to learn from many uncorrelated experiences, thus reducing the impact on-policy training has on local optimal convergence. Thus, is seen as a replacement for an experience replay buffer. The greatest benefit of the A3C and other asynchronous method is the reduced training time over single actor algorithms, especially when using distributed computing \cite{impala}. \\ 

\noindent A2C \cite{ACKTR} is a synchronous version of A3C, where multiple actors interact synchronously using a single shared policy-network. Running these environments synchronously removes the need to correct for any off-policy correction and allows for greater GPU utilisation.

\begin{figure}[H]
	\begin{tcolorbox}[title = $n$-step Synchronous Advantage Actor-Critic (A2C) Algorithm]
		\begin{algorithmic}
			\State{Randomly initialize policy $\pi$ and value function $V$ with parameters $\theta$, $\theta_v$ respectively}
			\State{Create $N_e$ environments}
			\State{$n$ - number of steps to calculate TD over}
			\State{$R^n$ - $n$-step return }
			\Repeat
			\For {actor $1 \hdots N_e$}
			\For {time $t\hdots t+n$}
			\State{Sample actions $a_t \sim \pi(a|s_t,\theta)$}
			\State{Execute actions $a_t$ and observe rewards $r_t$ and next states $s_{t+1}$}
			\EndFor
			\EndFor \\
			\State{$R^n_t= \begin{cases}
				0 & {\text { for terminal } s_{t}} \\
				V\left(s_{t} | \theta_v \right) & {\text { for non-terminal } s_{t} \qquad \text { Bootstrap from last state }}
				\end{cases}$} \\
			\For{$t+n,\, t+n-1,\, ... t$} \\
			\State{$R^n_t = \begin{cases}
				r_t & {\text { for terminal } s_{t}} \\
				r_t + \gamma R^n_{t+1} & {\text { for non-terminal } s_{t} }
				\end{cases} $}
			\EndFor
			\State{$ \nabla_\theta L(\theta) = \mathbb{E}_{\pi, \forall e \in N_e} \left[\nabla_{\theta} \log \pi(a|s,\theta)(R^n-V(s|\theta_v)
				+ \beta_{H} \nabla_{\theta}H(\pi(s,\theta)) \right]$} \\
			\State{$\nabla_{\theta_v} L(\theta_v) = \mathbb{E}_{\pi, \forall e \in N_e} \left[\frac{\partial(R^n-V(s|\theta_v))^{2}}{\partial \theta_v} \right]$}
			\Until{$t_{max}$}
		\end{algorithmic}
		
	\end{tcolorbox}
	\caption[Advantage Actor-Critic (A2C) algorithm]{A2C algorithm adapted from  \cite{Async} and \cite{baselines}} \label{fig:A2C}	
\end{figure}

\subsection{Synchronous DDQN}
In \cite{Async} the asynchronous multi-actor framework was also extended to include versions which use state-action value functions, e.g. SARSA and Q-Learning. Synchronous $n$-step Double DQN (Sync DDQN), is a synchronous version of the async DQN algorithm proposed in \cite{Async}, with Double DQN parametrisation following \cite{DDQN}. The network used to provide the value of action was a fixed copy of the online network $\theta$. The online network was used to select the action $a_t$ under the epsilon greedy policy. The value used to bootstrap the estimated future return from \ref{fig:A2C} is then replaced by $Q(s_t, \arg\max_a Q(s_t,a|\theta_i)\,\,|\theta_{i-1})$, where $\theta_{i-1}$ is a periodic copy of $\theta$ updated every $C$ steps similar to \ref{fig:DQN}.

\subsection{Proximal Policy Optimisation}
\subsubsection{Trust Regions}
In reinforcement learning the training data is dynamically produced by the agent. This means too large policy update can significantly alter the behaviour of the agent. Too large an update can cause the agent to move into parameter space where it converges on sub-optimal behaviour or worse, it no longer receives any rewards from the environment. A safer method to update the policy is to stick within a trusted region of the current policy as to not cause too dramatic updates. In Trust Region Policy Optimisation TPRO, \cite{TRPO} the surrogate objective function for a specific state, action pair $s_t,a_t$ is the policy scaled by the policy of this pair w.r.t. the old policy parameters $\theta_{old}$, penalised by the KL-divergence between the old policy and the new policy. This is given by the equation 
\begin{equation}
\underset{\theta}{\operatorname{maximize}}\,\, \mathbb{E}_{t}\left[\frac{\pi \left(a_{t} | s_{t}, \theta \right)}{\pi \left(a_{t} | s_{t}, \theta_{old} \right)} {A}_{t}-\beta \operatorname{KL}\left(\pi \left( \cdot | s_{t}, \theta_{old} \right) || \, \pi \left(\cdot | s_{t}, \theta \right) \right)\right] \cite{PPO}.
\end{equation}

$\beta$ is a constant that controls the strength of the KL-divergence penalty. The original derived quantity leads to a too small an update \cite{TRPO}, thus requires tuning for each task, or requires further complexity by adaptively changing the strength throughout training \cite{PPO}. In \cite{TRPO} they overcome this by optimising the surrogate objective $\mathbb{E}_{t}\left[\frac{\pi \left(a_{t} | s_{t}, \theta \right)}{\pi \left(a_{t} | s_{t}, \theta_{old} \right)} {A}_{t} \right]$
under the constraint that the KL-divergence is less than some scalar value $\delta$, instead of being penalised by it. This requires second order optimisation methods introducing even further complexity.\\

\subsubsection{PPO}
\noindent Proximal Policy Optimisation PPO \cite{PPO}, is a first order optimisation method that is used to calculate a lower-bound for the policy update. This is done in the form of clipping the policy ratio. The minimum value between the unclipped surrogate policy objective, and the clipped surrogate policy objective is chosen to perform the update on $\theta$. 
\begin{equation}
\underset{\theta}{\operatorname{maximize}}\,\, \mathbb{E}_{t}\left[ min\left(\frac{\pi(a_{t} | s_{t}, \theta)}{\pi(a_{t} | s_{t}, \theta_{old})} {A}_{t}, \,\, 
clip \left( \frac{\pi(a_{t} | s_{t}, {\theta})}{\pi(a_{t} | s_{t}, \theta_{old})}, \, 1-\epsilon, \, 1+\epsilon \right) {A}_{t}  \right) \right] \cite{PPO}.
\end{equation}
PPO is optimised over several epochs for each $n$-step sample, no theoretical justification is given in \cite{PPO} for this. This is a problem for policy gradient methods as after the first epoch the updated parameters are longer the parameters that produced the current training data, and thus further epochs are off-policy updates. The use of multiple updates could be justified by claiming the parameter updates are small enough as they constrained by the old policy and the clipping bound, such that the new policy is never too far from the old policy. Thus, reducing the chances of sub-optimal convergence. \\

\begin{figure}[H]
	\begin{tcolorbox}[title = Proximal Policy Optimisation (PPO) Algorithm]
		\begin{algorithmic}
			\State{Randomly initialize policy $\pi$ and value function $V$ with parameters $\theta$, $\theta_v$ respectively}
			\State{Create $N_e$ environments}
			\State{$n$ - number of steps to calculate TD over}
			\State{$\theta_{old} \leftarrow \theta$}
			\Repeat
			\For {actor $1 \hdots N_e$}
			\For {time $t\hdots t+n$}
			\State{Sample actions $a_t \sim \pi(a|s_t,\theta)$}
			\State{Execute actions $a_t$ and observe rewards $r_t$ and next states $s_{t+1}$}
			\EndFor
			\EndFor
			\State{Compute Advantages $A$ and Returns $R$}
			\For{epoch $1,\hdots,K$}
			\For{minibatch $1,\hdots,N$}
			\State{sample minibatch transitions randomly without replacement $(s_j,a_j,A_j,R_j)$ }
			\setstretch{1.5}{
				\State{$r(\theta)  = \frac{\pi(a_j | s_j, \theta)}{\pi(a_j | s_j, \theta_{old})}$}
				\State{$\nabla_\theta L(\theta) = \mathbb{E}\left[ \nabla_{\theta} \,\, min\left( r(\theta) {A}_j, \,\, 
					clip \left( r(\theta), \, 1-\epsilon, \, 1+\epsilon \right) {A}_j  \right)
					+ \beta_{H} \nabla_{\theta}H(\pi(s,\theta)) \right]$} 
				\State{$\nabla_{\theta_v} L(\theta_v) = \mathbb{E}_{\pi} \left[\frac{\partial(R^n-V(s|\theta_v))^{2}}{\partial \theta_v} \right]$}
			}
			\EndFor
			\EndFor
			\State{$\theta_{old} \leftarrow \theta$}
			\Until{$t_{max}$}
		\end{algorithmic}
		
	\end{tcolorbox}
	\caption[Proximal Policy Optimisation (PPO) algorithm]{Proximal Policy Optimisation algorithm adapted from \cite{PPO} } \label{fig:PPO}	
\end{figure}

\noindent As stated previously the Q Learning algorithm suffers from a maximisation bias and poor exploration. To improve on \cite{DQN}, methods such as $TD(n)$ and Double Q Learning as previously explained can be implemented along with additional reinforcement learning methods such as Duelling Q Networks, A3C and prioritised experience replay. These provide significant gains on the initial DQN algorithm as shown in \cite{Async} and \cite{Rainbow}.
Although, even after combining all these deep reinforcement learning methods, deep reinforcement learning still suffers from sampling inefficiency requiring up to 50 million state samples (200 million frames) in order to to converge on an policy equivalent to human-level performance in Atari-2600 games. Furthermore, all of these improvements fail to achieve near human level performance in sparse extrinsic reward games such as MontezumaRevenge or PrivateEye. 200 million frames is equivalent to playing $\sim$ 38 days in real-time, playing at 60fps. As stated earlier this is extremely inefficient learning when contrasted against humans who are often able to perform well in both dense and sparse reward games after a single or few attempts.

\section{Deep Reinforcement Solutions to Sparse Extrinsic Rewards}
\subsection{Curiosity Driven Exploration via Next-State Prediction}
Curiosity driven exploration in deep reinforcement learning as seen in \cite{Curiosity} attempts to solve the lack of frequent rewards by introducing a intrinsic reward function. The intrinsic reward function provides rewards when the agent experiences novel states. In order to avoid the agent from seeking states of high stochasticity which the agent cannot model e.g. leaves fluttering in a breeze \cite{Curiosity}, the agent learns an exploration model only for features that the agent can interact with.
The intrinsic reward is calculated by intrinsic curiosity model (ICM) in \cite{Curiosity}. The ICM consists of two models the inverse dynamics model which tries to predict the action $a_t$ given an encoded version of states $\phi(s_{t})$ and $\phi(s_{t+1})$ and the forward model that tries to predict the next encoded state $\phi(s_{t+1})$ using the actual action taken by the agent $a_t$ and the encoded state $\phi(s_t)$. In \cite{Curiosity_Scale} the intrinsic reward $i_t$ is given by the mean squared error $i_t = |\hat{\phi}(s_t) - \phi(s_{t+1})|^2$. By jointly optimising the inverse and forward model, the agent gains no benefit of trying to model any part of the environment that isn't affected by the action taken \cite{Curiosity}. However, this introduces a problem as parts of the environment that are not immediately affected by the agent's policy, may be important in the future \cite{Curiosity_Scale}.

\subsection{Unsupervised Auxiliary Tasks}
The UNsupervised REinforcement and Auxiliary Learning (UNREAL) \cite{Auxiliary} agent increases the sample efficiency of the A3C agent by extracting additional reward signals from the environment.

\begin{figure}[h]
	\centering
	\includegraphics[width=0.95\linewidth]{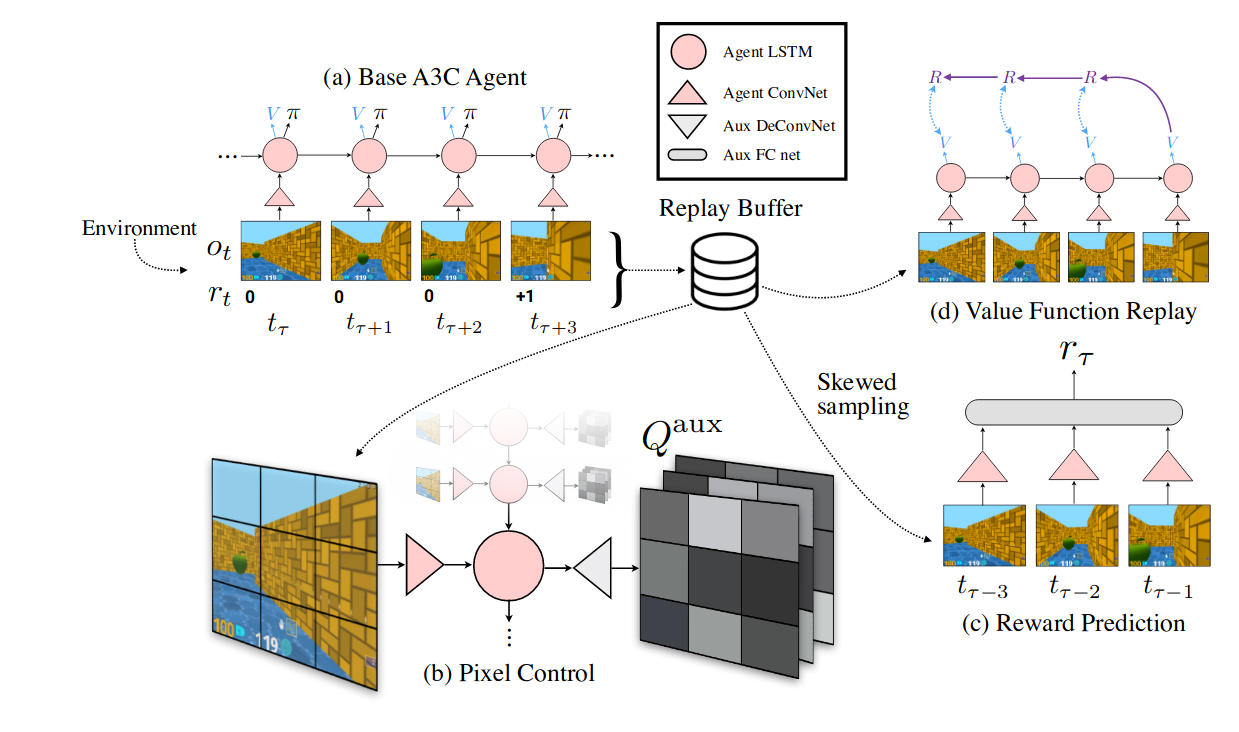}
	\caption[Overview of the UNREAL agent]{Overview of the UNREAL agent, image taken from \cite{Auxiliary} with authorial permission. This figure shows the auxiliary tasks (b) Pixel Control, (c) Reward Prediction and (d) Value Function Replay for the UNREAL agent.} \label{fig:UNREAL}
\end{figure}

One additional task is that of pixel control (PC), where an auxiliary policy $Q^{aux}$ is trained to maximise the pixel intensity across different sections of the image input over an $n$-step period as seen in Figure \ref{fig:UNREAL}(b). The auxiliary state-action function is trained by minimising the mean-squared error of the between the predicted $n$-step pixel change given as
\begin{equation}
L_{Q^{aux}} = \mathbb{E}\left[\left(R^n_{t}+\gamma^{n} \max _{a^{\prime}} Q^{aux}\left(s^{\prime}, a^{\prime}| \theta_{i-1}\right)-Q^{aux}(s, a| \theta_i)\right)^{2}\right]  \cite{Auxiliary}.
\end{equation}
The other auxiliary tasks are that of reward prediction (RP), where the agent tries to predict the reward received only for the next step given three preceding frames Figure \ref{fig:UNREAL}(c), and finally the value function replay (VR) which further trains the value function $V(s_t)$ from a replay buffer \ref{fig:UNREAL}(d). All three auxiliary tasks are trained off-policy where input state sequences are sampled from a small experience replay buffer. As the sequences sampled have different starting points to the online rollouts, this allows the agent to learn new temporal relations between states \cite{Auxiliary}. Here the three auxiliary tasks use the agents' existing neural network to jointly optimise the network for all tasks. The total loss function is then, 
\begin{equation}
L = L_{A3C} + \lambda_{PC} L_{Q^{aux}} + \lambda_{VR} L_{VR} + \lambda_{RP} L_{RP}
\end{equation} The auxiliary tasks do not directly affect the policy $\pi$, but rather shape the features in the encoder network which could then affect the agent's behaviour. However, Unlike the curiosity driven exploration the pixel control task learns to maximise the expected change in pixels, this could possibly indirectly lead to the agent seeking states of high stochasticity if $L_{Q^{aux}}$ is too large, as it will bias the encoded features towards those that change the pixel intensity the most. 

\subsection{Random Network Distillation}
The novelty of a state can be also measured by the number of counts the agent has been in that state, in \cite{unifying-count-based} this is combined with intrinsic motivation by providing an additional reward signal which is inversely proportional to the count. This method was simplified in \cite{RND} via using a Random Network Distillation (RND). RND consists of a base policy $\pi_\theta$, as well as randomly initialised target and predictor networks $f$ and  $\hat{f}$ respectively. The target network is fixed after initialisation and the predictor network is trained to predict the output of the target network for the next state $s_{t+1}$. The intrinsic reward is defined as the mean squared error between the two outputs $i_t = |\hat{f}(s_{t+1}) - f(s_{t+1})|^2$ and the predictor network is trained by minimising this loss \cite{RND}. By using a lower dimensional representation this should reduce the effect of the agent seeking states of high stochasticity as mentioned earlier. \\

\noindent Minimising the distance between $f$ and $\hat{f}$ implicitly learns the count of the number times a state has been visited, as the predictor network cannot predict the target network output with greater than random probability without first visiting the state. As the target network $f$ output is deterministic i.e. constant, this means the output is learnable and the loss can approach at least a local minima through gradient descent. This is notably different to a forward dynamics model, where for certain non-stochastic environments such as  simplistic physics based environments the next state can be predicted without necessarily having to have explicitly visited that state, as the forward dynamics model learns the true transition model $\mathcal{P}^{a}_{ss'}$. This is likened to the agent getting bored in \cite{Curiosity}, as the outcome becomes more predictable hence less novel. 

\subsection{Hindsight Experience Replay}
Auxiliary Tasks and curiosity driven exploration both attempt to explore unknown states by maximising the statistical novelty of a state compared to states the agent previously experienced. This only will work if there is a variation amongst the states being explored, however in some tasks the states lack diversity hence the ICM and pixel control policies will not sufficiently be able to explore the state space \cite{HER}. This is especially crucial for large state spaces with very few rewards. The proposed solution uses every episode the agent performs to provide a reward even if the agent does not achieve the said goal. This is done by creating virtual goals which use the final the state the agent ended up in as the desired outcome. An example given in \cite{HER} is using a robotic arm to moving an object to a specific coordinate (x,y). For a large x,y plane randomly sampling a specific location would be statistically negligible, so after every episode the agent updates it's value function by "pretending" that the final (x,y) coordinate was the desired outcome. Hence, the agent learns to move the object to any specified coordinate.

\chapter{Evaluation and Testing} \label{chapter5}
\section{Environments}
In order to test the efficacy of the algorithms discussed above, a suitable MDP environment must be used. Several environments were used to provide a variety of environment complexities and task difficulties, but crucially a difference in sparsity of the extrinsic reward received. The environments include that of classic control games \{CartPole, Acrobot, MountainCar\} and a small selection of Atari games \{SpaceInvaders, Freeway, MontezumaRevenge\} were chosen to meet the above requirements, as well as their popularity so that there were plenty of reference results in the literature. All environments were executed using OpenAI's gym toolkit.

\subsection{Classic Control}
These simplistic environments represent the state of the environment with a vector $ s \in \mathbb{R}^n $, with each feature corresponding to a physical variable of the system that the agent needs to learn to control in order to gain rewards. The CartPole-v1 game consists of an unstable inverted pendulum (pole) attached to a cart and the goal is to keep the pole upright. The agent can move the cart left or right via applying a discrete force of $\pm 1$ along the horizontal axis of the cart. The game finishes when the pole moves greater than 15\textdegree either side of the upright position, moving $\pm 2.4$ units from the centre or after 500 steps. The agent receives a reward +1 for each step until the episode is finished \cite{gym-CartPole-v1}. \\ 

\noindent The Acrobot-v1 environment consists of a double pendulum in which the goal is to try to swing the tip of the lower pendulum to a specific target height. The agent can move the pendulum by applying a torque of \{+1, 0, -1\} to the joining link. Every step the agent receives a negative reward other than the target state \cite{gym-Acrobot-v1}. \\ 

\noindent The final control task is MountainCar-v0, where the goal is to reach the top of the mountain from the bottom of the valley. However, the car does not have enough power to directly drive up the side, therefore in order to solve the problem, the agent must gain momentum from oscillating either up slide of the mountain. The agent receives a -1 reward for every step the agent is not at the top and the episode terminates at 200 steps. Therefore the agent will receive a -200 reward unless it learns to reach the top in under 200 moves \cite{gym-MountainCar-v0}. \\

\noindent At first glance these as all these tasks contain a dense extrinsic reward signal as they produce a non neutral at each step, so one might expect that they should be solved easily. However, under further inspection in Acrobot and MountainCar tasks, the rewards contain very little information about the state-action value, as all actions are considered equally bad regardless of how close the agent is to the optimal behaviour. An agent must sufficiently explore the state space in order to achieve the a higher reward, hence learn from it. 

\subsection{Atari 2600} 
For the Atari 2600 games, OpenAI provides wrapper to the Arcade Learning Environment (ALE). The ALE produces a 210x160x3 RGB image for each frame, however many frames are duplicates or are very similar so it has been a common practise since \cite{DQN} to only consider the nth successive frames, repeating the same action n times. This helps reduce the computational cost by concatenating n steps in the MDP onto a single step. Another preprocessing step performed is to convert the images to greyscale and scale and crop the image to a size 84x84. The final preprocessing step for the Atari environments is to concatenate the current frame with the previous k frames into a single image. This solves the partially observable MDP problem, providing additional temporal information into the state (such as velocity of the ball in pong) that would not be present in a static frame. This allows the use of convolutional neural networks to be used as the non-linear state-value function approximator.

As the environment is deterministic a random amount of `no-op'  (neutral) actions were taken at the start of each episode, this changes the initial starting state, and is done to prevent the agent `memorising' good actions from frequently visited target states. Sometimes an loss of life, done or terminal flag is used to increase the agent's learning efficiency and, performance. This means when bootstrapping for the estimate of the expected future return, a loss of life is treated as the terminal state. The reasoning being that using the flag will allow the agent to quickly learn losing a life is to be avoided. Another commonly used practise is to cap the maximum number of steps an agent can take in the environment before the episode terminates. This penalises idle behaviour, thus should make the agent more efficient. The exact environment parameters are given in \ref{Testing} \\ 

In SpaceInvaders a series of waves of aliens zig-zag down the screen firing at the agent, the agent's goal is to shoot the aliens before they reach a certain threshold height. The game increases in difficulty as the aliens increase their speed of descent as their numbers go down. a life is lost when the aliens reach the threshold and the game is ended when all 3 lives are lost or the player has defeated all waves of aliens. \\ 

\noindent In the game of Freeway the goal is for the player to reach the other side of the road without being hit by the oncoming traffic. The player receives a reward of +1 when reaching the other side and 0 otherwise.
In the game MontezumaRevenge the goal is to explore temple rooms and collect amulets. Navigating the different rooms requires the collecting objects such as keys found throughout the level, as well as stringing together a series of complex movements to avoid falling and traps.\\

\noindent The games were chosen for varying sparsity of rewards as according to \cite{unifying-count-based}, SpaceInvaders provide a human-optimal environment with relatively dense reward feedback and Freeway and MontezumaRevenge provide the sparse reward environments with easy and difficult search respectively.

\chapter{Implementation and Analysis} \label{chapter6}

\section{Software and Hardware}
Python was the language of choice for implementing these algorithms as it is widely used in the reinforcement learning community, such as OpenAI's baselines \cite{baselines} allowing for better troubleshooting and debugging. Also, Python is one the most common used API for neural network packages, such as the popular Tensorflow and Pytorch packages. 
All neural networks were implemented using Tensorflow's python API. Tensorflow is a Machine Learning framework that allows users to build graphical models using `tensors', Tensorflow's internal representation of a generalised matrix \footnote{Note that these tensors are distinct in notation and properties from the tensors in tensor calculus \cite{tensor}}, and provides built-in automatic differentiation, which automatically handles gradient calculation and backpropagation. And all the experiments were ran on either a single machine with a 8-core, 16-thread AMD Ryzen\textsuperscript{\tiny TM} 1800x with a single NVIDIA\textsuperscript{\tiny\textregistered} GeForce\textsuperscript{\tiny\textregistered} RTX 2070 GPU and 48GB of DDR4 RAM, or, on a single GPU-node on the  Sheffield Advanced Research Computer (ShARC) high performance computing system, which contained a NVIDIA\textsuperscript{\tiny\textregistered} Tesla K80 with a Intel\textsuperscript{\tiny\textregistered} Xeon\textsuperscript{\tiny TM} E5-2630 v3 CPU. All references to computational speed and timing are those observed using the single machine setup. 

\section{Scalable Reinforcement Learning}
In order to perform the Atari experiments over 200 million frames, it is crucial that the implementation use multiple workers to run the environments,  these methods scale number of steps taken in the environment per second sub linearly. Many different scalable frameworks were attempted such as A3C and IMPALA \cite{impala}, however due to Tensorflow's incompatibility with python's multiprocessing library these methods required multithreading, or using Tensorflow's distributed training framework. Due python's Global Interpretable Lock, multithreading is not truly parallel, only concurrent and attempted implementations using Tensorflow's distributed training for a single machine were relatively slow compared to results shown in the literature \cite{impala}. The A2C framework, however was observed to increase the number of steps per second by a factor of 10, from $\sim$ 250 to $\sim$ 2300-2700 steps per second. The A2C framework reduces the need or multiple copies of the network by running a single step in multiple environment in parallel. The implementation of running multiple environments synchronously in parallel was adapted from OpenAI's implementation \cite{baselines}, and achieved this by running an individual environment in a separate process using python's multiprocessing library. Having the environments run synchronously reduces the overhead of passing gradients between processes as well as the need to account for off-policy corrections. However, the most crucial benefit is the increased efficiency of the use GPU when using a single machine. This is as value estimations can be ran in a single batch decreasing the time spent of transferring data onto the GPU, which is a large bottleneck for GPU applications. 

\section{Encoder Neural Networks}
All implementations used neural networks to encode the state $s_t$ onto a high dimensional feature vector, which then would be used to predict the value function of policy distribution from. In order to be consistent across experiments, every agent used the same encoder networks architectures. \\

\noindent For the Classic Control tasks the encoder network was two MLP layers of with 64 hidden units each, under the ReLU activation function. \\

\noindent For the Atari experiments a CNN encoder network was used, as in \cite{nature}, as it is a commonly used to benchmark to test different algorithms on. This CNN is parametrised as following, first a 2D convolutional layer with filter size of $\left[8,8\right]$ of with 32 output channels and stride $\left[4,4\right]$, following a 2D convolutional layer with filter size $\left[4,4\right]$, 64 output channels and stride $\left[2,2\right]$. A final convolutional layer with 64 output channels, filter size $\left[3,3\right]$ and stride $\left[1,1\right]$. The resulting convolutional layer was flattened to a vector and a final fully connected MLP layer with 512 hidden units. All layers were under the ReLU activation function, and all convolutional layers had the 'VALID' padding. The observational input range for the policy CNN encoder is scaled from $[0,255]$ to $[0,1]$ for each pixel by diving by 255 to avoid to large ReLU activations as it is unbounded in the positive direction equation \ref{eq:ReLU}. \\

\noindent Both encoder networks use the Glorot uniform weight initialisation and zero bias initialisation.

\section{A2C}
For the Advantage-Actor Critic A2C network, The actor and the critic share the same encoder network with the value function and unnormalised log probabilities (logits) being linear projections (via a fully connected MLP layer with no activation) from the encoded feature vector. To get the policy distribution $\pi$, the softmax activation function was applied to the logits, resulting in a policy distribution over the discrete set of available actions. The gradient of the entropy over the policy distribution $\beta_H \nabla_\theta H(\pi(s,\theta))$ is added to the loss function following \cite{Async} in order to provide regularisation of the policy such that is doesn't converge on a sub optimal policy. By maximising the entropy, the uniformity of the probability distribution increases, this should increases exploration as previously unlikely actions are sampled with greater frequency. The strength of the entropy maximisation is controlled by the parameter $\beta_H$ \cite{Async}.

\section{Synchronous DDQN}
For the Synchronous $n$-step DDQN network, the state-action value function was a linear projection from the encoder network. Here double Q-learning was used to reduce maximisation bias.

\section{PPO}
The PPO implementation uses the exact same model as the A2C model mentioned above. A separate network for the old policy parameters $\theta_{old}$ is not used, instead the policies for each rollout are stored and then passed into the network during backpropagation.

\section{Unsupervised Auxiliary tasks}
\subsection{Original Implementation}
\subsubsection{A3C Policy}
In \cite{Auxiliary} the A3C agent is parametrised by an encoded CNN-LSTM network, which takes in an RGB colour image of size $\left[84,84\right]$ as it's input state $s_t$ for time t. It then encodes this onto a feature vector using a small two layer CNN following the original DQN paper implementation \cite{DQN}. The CNN-encoded state is concatenated with the previous reward and action is then fed into a LSTM layer with 256 hidden units, from this the policy and value function are projected as described in the A2C implementation. 

\subsubsection{Pixel Control}
For The pixel reward for each time step the image at time $t$ is cropped to a the central region of size $80\times80$ pixels, and subdivided into $20\times20$ cells of size $4\times4$. The pixel reward for time $t$ is then defined by the absolute difference between the mean over all colour channels and all pixels for each cell, of the input image $s_t$ and the previous image $s_{t-1}$.

In order to produce the $Q^{aux}$ estimate of expected pixel change from each subdivided cell region, a MLP layer with relu is added the LSTM output layer, this is then reshaped to size $32\times7\times7$. A deconvolutional layer with 32 output channels, filter size $\left[3,3\right]$ and stride one produces a $32\time9\times9$ feature map. Two deconvolutional layers with output channels 1, number of actions are applied to the $9\times9$ feature map concurrently, both with filter size $\left[2,2\right]$. This produces a value function and advantage function estimate which is used to produce the $Q^{aux}$ estimate via duelling neural networks given by \cite{Auxiliary}
\begin{equation}
Q^{aux}(s,a|\theta) = ReLU \left(V(s|\theta) + A(s,a|\theta) -  \frac{1}{|\mathcal{A}|}\sum_{a'\in \mathcal{A}}A(s,a'|\theta)\right) \,\,\cite{Dueling_Q}.
\end{equation}

\subsubsection{Reward Prediction}
For the reward prediction a reward $r_t$ is sampled from the experience replay in a skewed manner such that the probability mass for non-zero rewards is equal 0.5. The three proceeding images $\left[ s_{t-2},s_{t-1},s_{t}\right]$ \footnote{as $r_t$ is reward from $s_t\rightarrow s_{t+1}$} are passed into the CNN encoder, the flattened output features across all three images are concatenated into a single feature vector. This is then passed into a MLP layer with 128 hidden units with the ReLU activation function. This is then projected via another MLP layer onto a categorical distribution using the softmax function. The categories of this distribution are positive negative and neutral rewards, which should be easier to learn than the exact value of $r_t$, this gradient is calculated via cross entropy loss. 

\subsubsection{Value Replay}
A small experience replay buffer stores the most recent 2000 states for each actor, the gradient of the value function replay task is calculated via mean squared error in the same fashion as the base A3C agent. \\

\subsection{Custom A2C-CNN implementation (UNREAL-A2C2)}
\subsubsection{A2C Policy}
In order to be a more accurate comparison to the base A2C agent and for greater GPU utilisation, an CNN-only A2C version of \cite{Auxiliary} is implemented. In order to get this to work the following changes were made, first the states are converted to greyscale and stacked to the match base A2C and Sync-DDQN implementations. In order to be more comparable to the A2C experiments the action-reward concatenation to the encoded state before the value and policy was not performed, as this provides extra temporal information that may improve performance.

\subsubsection{Pixel Control}
The pixel control reward at each time step is then altered to be defined as the absolute mean  difference across the greyscale subdivided region of the most recent image $x_t$ in the concatenated state $s_t = \left[ x_{t-3},x_{t-2},x_{t-1},x_t \right]$
\begin{equation}
r^{PC}_t = |subdivided(x_t) - subdivided(x_{t-1})|. \nonumber
\end{equation}
Other changes, made to the pixel control task were to not perform the cropping operation and subdivide the image into $21\times21$ cells. The only change this requires to the deconvolutional network is to increase the number of size of the MLP layer before the first feat map from size $32\times7\times7$ to $32\times8\times8$. Finally the greyscale images were normalised between the values of (0,1) using a simple min-max normalisation \cite{feat-norm} in order to avoid manually fitting the $\lambda_{PC}$ term per environment. However, this does introduce a small non-stationarity affect into calculating the pixel rewards as the maximum pixel value of the dataset may increase over the course of training. This is less pronounced for than it would be for per-channel RGB min max normalisation, as well as being more stationary that per-pixel z-score image normalisation. This is because in episodic video games tasks many start in a specific location, and when progressing to different areas often means a change in environment textures, thus possibly changing the normalisation parameters significantly. Whereas the change in maximum pixel intensity for a greyscaled images is expected to be negligible. 

\subsubsection{Reward Prediction}
The reward prediction task is the same as the original implementation as described above but, without the frame concatenation as the state already provides necessary temporal information. Instead of sampling rewards from all workers, the worker with the highest reward is greedily chosen to reduce the chance of sparse rewards going unmissed, as well as increasing computational efficiency. \\

\noindent Because of the noticeable modifications to the original UNREAL agent's algorithm, to distinguish between the two algorithms, the custom agent used is hence referred to as the UNREAL A2C-CNN or UNREAL-A2C2 agent.

\section{RND}
\subsection{Intrinsic Reward and Value Function}
Following \cite{RND}, The base policy is a PPO policy with an additional separate critic is used to calculate the value of the intrinsic rewards $V_i(s)$. The weighted combination of the extrinsic and intrinsic advantages is used to update the policy, and the discount factor for the extrinsic reward is increased to $0.999$ as to match \cite{RND}. The intrinsic rewards are non-episodic, as \cite{Curiosity_Scale} argues that this increases exploration, and they are normalised so that the scale is independent of the observational space \cite{RND}. This is achieved by dividing the intrinsic reward by the running estimate of the standard deviation of the intrinsic return. However the intrinsic rewards are non-episodic so the true return cannot be calculated as it goes on indefinitely. It is also non-stationary as it is produced by a changing parametrised function, so using the intrinsic value estimate to approximate the return is suspected to be an unstable normalisation technique. So the normalisation used in \cite{RND_code} is used, where the intrinsic reward is normalised by the standard deviation of the inversely discount rewards as the past rewards are calculable. This can be explicitly defined as,
\begin{equation*}
i_t = \frac{\left|f(s_{t+1}) - \hat{f}(s_{t+1}) \right|^2 }{\displaystyle \sigma\left[\sum_{k=0}^t \gamma^{t-k}i_t \right]}.
\end{equation*}

\noindent In practise the running estimate of the standard deviation is updated after every $n$-step rollout and intrinsic return is calculated online.  

\subsection{Target and Predictor networks} \label{f,f}
As the target network is fixed, it cannot adjust to the scale of the observational inputs, thus normalisation techniques are required in order ensure the output scale is consistent across different environments \cite{RND}. The input to the predictor and target networks is the final image $x_t$ in the state $s_t = [x_{t-3}, x_{t-2}, x_{t-1}, x_t]$, this is then normalised by subtracting the running estimate of the per-pixel mean and dividing it by the running estimate of the per-pixel standard deviation and finally being clipped between the range -5, 5. However, This introduces non-stationarity to target network output thus further non-stationarity to the intrinsic reward. The running estimates of the observational mean and standard deviations are initialised by running a random agent for 50 rollouts for each actor and continually updated during training.\\

\noindent The target and predictor network are both separate models from the policy as to ensure bias free reward prediction, they both use the same convolutional encoder architecture as the policy network.

The target network has a final MLP layer of size 512 with no activation after the encoder. For the predictor network after the convolutional encoder it has two successive MLP layers of size 512, both with the ReLU activation followed by a final MLP layer with size 512 with no activation. The target and predictor networks use a leaky ReLU activations for the convolutional layers, as well as all layers being initialised with orthogonal initialisation with gain $\sqrt{2}$. This was not done to increase performance but to ensure the output of the target and prediction networks were of the same magnitude as the \cite{RND_code} implementation, so that the hyperparameters taken from \cite{RND} are valid.

\section{ICM}
For curious agents using the next-state prediction, the ICM model is used to provide the additional reward for the PPO policy with separate intrinsic value function. The ICM is parametrised  by additional separate encoder network using the same architecture and weight initialisation as the base policy encoder, but replacing the observational scaling with the observational normalisation as described in the RND implementation. The output from the encoder $\phi(s_t)$ is concatenated the with action $a_t$ and fed into the forward model. The forward model consists of one fully connected MLP layer with ReLU activation, with the number of hidden units the same size of the encoded state, i.e. 512 and 64 for the CNN and MLP encoder respectively. A following MLP with no activation and same size of the previous layer creates the predicted next state $\hat{\phi}(s_{t+1})$ \cite{Curiosity}. \\

\noindent For the inverse model the encoded state and next state $\phi(s_t)$ and $\phi(s_{t+1})$ are concatenated into a single feature vector and passed into a MLP layer with ReLU activation and size 512 and 64 for the CNN and MLP respectively. This is fed into a final MLP layer with no activation that maps onto the unnormalised logit probability distributions over the available actions producing $\hat{a_t}$. The inverse model is trained as a classification task using cross entropy between the predicted action $\hat{a_t}$ and actual action $a_t$ taken at time $t$, as well the forward model being trained via mean squared error $\left|\hat{\phi}(s_{t+1}) - \phi(s_{t+1}) \right|^2$ \cite{Curiosity}. Unlike the \cite{Curiosity_Scale} implementation the forward model and the inverse model gradients are both passed back to the encoder model as in the original implementation \cite{Curiosity}, with weighting $\beta L_{forward} + (1-\beta)L_{inverse}$, with $\beta = 0.2$.

\section{Random Network Distillation with Auxiliary Learning RANDAL}
RAndom Network Distillation with Auxiliary Learning (RANDAL) is a novel method of combining curiosity driven exploration with RND and auxiliary tasks was implemented as following. A base PPO policy with RND is combined with all auxiliary tasks as described in UNREAL-A2C2 implementation, in order to stop the intrinsic reward from decreasing too rapidly and thus exploring less, a separate intrinsic value function is used as in the RND agent. This allows the value replay be performed exclusively on the extrinsic value function, preserving the rate of change in $i_t$. As the intrinsic reward is non-stationary, the reward prediction is done only on the sign ($+, -, 0$) of the extrinsic reward, this is done in hopes to increase stability as extrinsic rewards are stationary as well as being able to continue to use the simplified categorical reward rather than the exact reward value. The target and predictor networks are identical to the RND implementation.

\section{Analysis}
For the baseline architectures of Sync-DDQN and A2C, it is expected that A2C agents shall outperform Sync-DDQN agents as it is a policy gradient method. As previously mentioned policy gradient methods provide are theorised to have more stable convergence properties, thus should find a more optimal than the value based Sync DDQN agents. However, the A2C algorithm is expected to perform worse on sparse reward tasks as it suffers for poor exploration, e.g. A2C will probably never learn to navigate MountainCar, Freeway or MontezumaRevenge. Again this is because it is an on-policy algorithm using the estimate of the optimal policy $\pi(a|s,\theta)$ to sample actions from each turn. The expectation of the Sync DDQN agents is that they should slightly underperform compared to the A2C for dense reward tasks but should perform better for tasks that benefit from random exploration, i.e. should outperform A2C on easy search sparse reward tasks, for the chosen environments this includes the Freeway, Acrobot and MountainCar tasks. \\

PPO agents should outperform A2C and Sync-DDQN agents on all tasks, as the multiple epoch minibatch training should ensure sparse rewards do not get lost in the gradient update, as well as the value function estimate converging to the true value of the policy quicker than the single update algorithms. This reiteration of the value function is comparable to the value function replay in the UNREAL architecture, thus the performance of the PPO agent should be closer to the UNREAL agents than the A2C and Sync-DDQN agents.

For the curious agents, the expectation is that the ICM and RND agents should perform equally well for high dimensional complex inputs, such as image observations in Atari games. This is because the forward model will not converge too quickly, providing similar amounts of rewards as the RND agents. For low dimensional non-complex environments, ICM agents are expected the magnitude of the intrinsic rewards is expected to decrease quickly as the observations are easily predictable, even for unseen states.

\chapter{Results and Discussion} \label{chapter7}

\section{Testing and Validation} \label{Testing}
\noindent For the Atari experiments the number of frames skipped (actions repeated) was 4, meaning each environment step is equivalent to 4 frames. The number of frames stacked per state was 4, the experiments were ran over 50 million steps, or 200 million frames as according to \cite{DQN} and the episode terminated after 4500 steps as to match \cite{RND}. The number of `no-op' actions was a random number between 0 and 7 steps, or 0-28 frames. The agent was validated every 1 million steps and the score was averaged over 50 episodes. During validation the episode was terminated after 10,000 steps approximately 11 minutes of real-time play to avoid games such as MontezumaRevenge carrying on indefinitely if the agent is stationary.\\

\noindent For the Classic Control experiments the ran over 2 million steps, the agent was validated following the same procedure as the Atari experiments except validated every 40,000 steps. \\

\section{Atari Experiments}
Due to the computational complexity and limited time of this project, no hyperparameter search was completed for the Atari domains, and the hyperparameters were taken from existing work in the literature. All agents use a gradients calculated with a clipping by normalisation value of 0.5 \cite{grad_clip} to help stabilise gradient updates. All figures show the mean score over 3 random starts, $\pm$ the standard deviation of the runs.
\subsection{Baselines}
The selected environments were all run for both the Synchronous $n$-step Double DQN and the A2C architectures as to provide a baseline to compare the more advanced algorithms to. As \cite{PPO} and \cite{RND} uses Adam optimisation, in order for the baselines A2C and Sync-DDQN to be a fairer comparison, Adam is used for both of these agents, with $\epsilon = 1\times 10^{-8}$ and $\beta_{m}=0.9, \,\, \beta_{v}=0.999$. Although, RMSProp is used in the original implementation \cite{Async}.\\ 
\begin{minipage}{\linewidth}
	\begin{minipage}{.45\linewidth}
		\begin{table}[H]
			\centering
			\begin{tabular}{ c | c }
				Hyperparameter & Value \\
				\hline
				Optimiser & Adam \\
				Learning rate & $1\times 10^{-3}$  \\
				Number of actors & 32  \\
				Entropy coefficient & $0.01^1$ \\
				Value coefficient & $0.5^2$ \\
				$n$-step period & $5^1$ \\ 
				Discount factor $\gamma$ & $0.99^1$ \\
				Gradient norm clipping & $0.5^2$ \\  
			\end{tabular}
			\caption[A2C Atari baseline hyperparameters]{A2C Atari baseline hyperparameters, taken from $^1$\cite{Async}, $^2$\cite{baselines}}
			\label{tbl:A2C Atari}
		\end{table}
	\end{minipage}
	\hfill
	\begin{minipage}{.45\linewidth}
		\begin{table}[H]
			\centering
			\begin{tabular}{ c | c }
				Hyperparameter & Value \\
				\hline
				Optimiser & Adam \\
				Learning rate & $1\times 10^{-3}$  \\
				Number of actors & 32  \\
				Target Network Period & $10,000$ steps $^3$\\
				Initial $\varepsilon$ & $1^1$ \\
				Final $\varepsilon$ & $0.01^1$ \\
				$\varepsilon_{test}$ & $0.01^1$ \\
				$n$-step period & $5^1$ \\
				Discount factor $\gamma$ & $0.99^1$ \\
				Gradient norm clipping & $0.5^2$ \\  
			\end{tabular}
			\caption[Synchronous Double DQN Atari baseline hyperparameters]{Synchronous Double DQN Atari baseline hyperparameters, taken from $^1$\cite{Async}, $^2$\cite{baselines}, $^3$\cite{DDQN}}
			\label{tbl:DDQN Atari}
		\end{table}
	\end{minipage}\\ \\
\end{minipage}
\\
\noindent In the Atari experiments for the value based Double DQN the exploration rate  $\varepsilon$ was linearly annealed over 2 million steps to provide sufficient exploration.

\begin{figure}[H]
	\centering
	\includegraphics[width=0.32\textwidth]{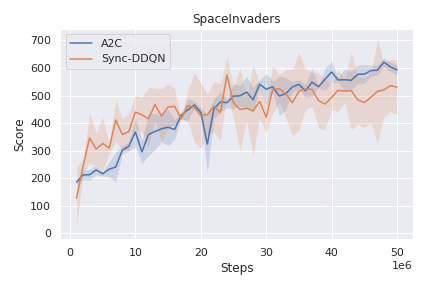}
	\includegraphics[width=0.32\textwidth]{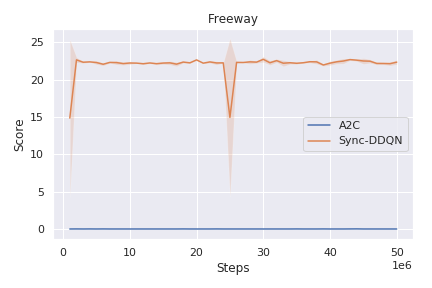}
	\includegraphics[width=0.32\textwidth]{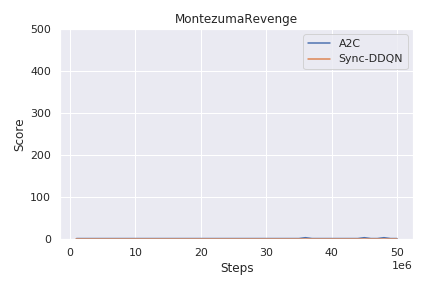}
	\caption[Atari baselines for A2C vs Sync-DDQN]{Atari Baselines, results show a mean validation score $\pm$ standard deviation over 3 random parameter initialisations run of the respective models for the number of steps in millions. The figure shows that the policy gradient A2C leads to better convergence for dense reward tasks however suffers from poor exploration and thus cannot learn to solve sparse reward tasks. This is contrasted against the Sync-DDQN which can solve the easy search sparse task Freeway with random search, however fails to learn from the more complex MontezumaRevenge sparse reward task.}
	\label{Atari_Baselines}
\end{figure}

\subsection{$\lambda$-Return and PPO} \label{sec:lambda}
In order to better connect the future rewards to the value of the current state $V(s_t)$, the $n$-step period was increased to 20 to match the \cite{Auxiliary} implementation, and to avoid having to manually fit the $n$-step period for each experiment the $\lambda$-return from equation \ref{eq:lambda} and GAE from equation \ref{eq:GAE} were used for the value-based and policy based methods respectively, with a value of $\lambda = 0.95$ taken from \cite{GAE}. For the PPO experiments, the $n$-step period was further increased to 128 to match the value used in \cite{Curiosity_Scale} and \cite{RND}. 

\begin{table}[H]
	\centering
	\begin{tabular}{ c | c }
		Hyperparameter & Value \\
		\hline
		Optimiser & Adam$^1$ \\
		Learning rate & $1\times 10^{-4}$  \\
		Number of actors & $32$  \\
		Entropy coefficient & $0.01^1$ \\
		Value coefficient & $0.5$ \\
		$n$-step period & $128^2$ \\ 
		Number of epochs & $4^2$ \\
		Number of minibatches & $4^2$ \\
		Discount factor $\gamma$ & $0.99^1$ \\
		PPO clip range & $\left[0.9,1.1\right]^2$ \\
		Gradient norm clipping & $0.5^3$ \\  
	\end{tabular}
	\caption[PPO Atari hyperparameters]{PPO Atari hyperparameters, taken from $^1$\cite{PPO}, $^2$\cite{RND} $^3$\cite{baselines}}
\end{table}

\begin{figure}[H]
	\centering
	\includegraphics[width=0.32\textwidth]{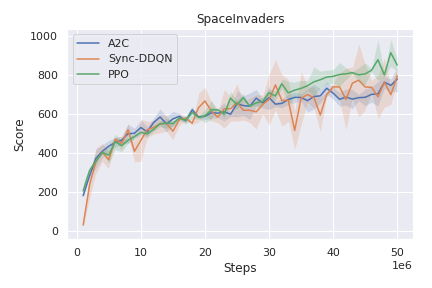}
	\includegraphics[width=0.32\textwidth]{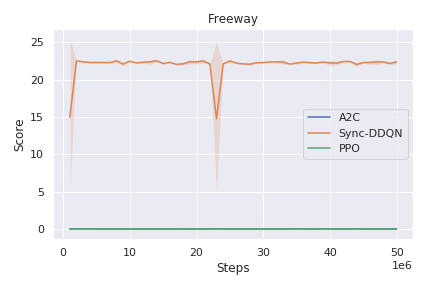}
	\includegraphics[width=0.32\textwidth]{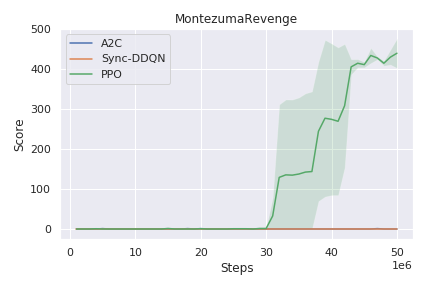}
	\caption[Atari experiments for PPO-GAE vs A2C-GAE and Sync-DDQN($\lambda$)]{Atari experiments for PPO with GAE vs A2C with GAE vs Sync-DDQN($\lambda$). The combination of increased $n$-step period and use of $\lambda$ return, significantly increase performance in the dense reward environment over the initial TD(5) return however, show no performance gains for the sparse reward tasks. Surprisingly the PPO agent does not outperform the A2C and Sync-DDQN agents on the easy search sparse reward task, where it was expected to outperform on all three tasks according to the results obtained in \cite{PPO}.}
	
	\label{lambda_Atari}
\end{figure}

\subsection{Sparse Reward Solutions}
The three sparse reward solutions being tested are, the UNREAL-A2C2 agent, the RND with PPO, and the novel RANDAL with PPO. All agents use the respective $n$-step period from their base policies in section \ref{sec:lambda} and all use GAE to improve performance. Due to the significantly slower algorithm implementation and time constraints of the project, a curiosity driven agent using the ICM was not tested for the Atari experiments.

\begin{minipage}{1\linewidth}
	\begin{minipage}{.45\linewidth}
		\centering
		\begin{tabular}{ c | c }
			Hyperparameter & Value \\
			\hline
			Optimiser & Adam$^1$ \\
			Learning rate & $1\times 10^{-4}\, ^{(1)}$  \\
			Number of actors & $32$  \\
			Entropy coefficient & $0.001^1$ \\
			Value coefficient & $0.5$ \\
			$n$-step period & $128^1$ \\ 
			Number of epochs & $4^1$ \\
			Number of minibatches & $4^1$ \\
			Extrinsic advantage coefficient & $2.0^1$ \\
			Intrinsic advantage coefficient &  $1.0^1$\\
			Extrinsic discount factor $\gamma_e$ & $0.999^1$ \\
			Intrinsic discount factor $\gamma_i$ & $0.99^1$ \\
			PPO clip range & $\left[0.9,1.1\right]^1$ \\
			Gradient norm clipping & $0.5$ \\  
		\end{tabular}
		\captionof{table}[RND Atari hyperparameters]{RND Atari hyperparameters, taken from $^1$\cite{RND}}
		\label{tbl:RND Atari}
	\end{minipage}
	\hfill
	\begin{minipage}{.45\linewidth}
		\centering
		\begin{tabular}{ c | c }
			Hyperparameter & Value \\
			\hline
			Optimiser & Adam \\
			Learning rate & $1\times 10^{-3}\,^{(1)}$  \\
			Number of actors & 32  \\
			Entropy coefficient & $0.001^1$ \\
			Value coefficient & $0.5^1$ \\
			$n$-step period & $20^1$ \\ 
			Discount factor $\gamma$ & $0.99^1$ \\
			Gradient norm clipping & $0.5$ \\  
			Reward prediction coefficient & $\,\,1^1$ \\
			Value replay coefficient & $\,\,1^1$ \\
			Pixel Control coefficient & $1\vphantom{^1}$ \\
			Replay length per actor & $2000^1$ \\
		\end{tabular}
		\captionof{table}[UNREAL-A2C2 Atari hyperparameters]{UNREAL-A2C2 Atari hyperparameters taken from $^1$\cite{Auxiliary}}
		\label{tbl:UNREAL Atari}
	\end{minipage}
\end{minipage}
\begin{minipage}{\linewidth}
	\begin{table}[H]
		\centering
		\begin{tabular}{ c | c }
			Hyperparameter & Value \\
			\hline
			Optimiser & Adam$^1$ \\
			Learning rate & $1\times 10^{-4}$  \\
			Number of actors & $32$  \\
			Entropy coefficient & $0.001$ \\
			Value coefficient & $0.5$ \\
			$n$-step period & $128$ \\ 
			Number of epochs & $4$ \\
			Number of minibatches & $4$ \\
			Extrinsic advantage coefficient & 2.0 \\
			Intrinsic advantage coefficient &  1.0\\
			Extrinsic discount factor $\gamma_e$ & $0.999$ \\
			Intrinsic discount factor $\gamma_i$ & $0.99$ \\
			PPO clip range & $\left[0.9,1.1\right]$ \\
			Gradient norm clipping & $0.5$ \\ 
			Reward prediction coefficient & $1$ \\
			Value replay coefficient & $1$ \\
			Pixel Control coefficient & $1$ \\
			Replay length per actor & $2000$ \\ 
		\end{tabular}
		\caption[RND Atari hyperparameters]{RANDAL Atari hyperparameters}
		\label{tbl:RANDAL Atari}
	\end{table}
\end{minipage}

\begin{figure}[H]
	\centering
	\includegraphics[width=0.32\textwidth]{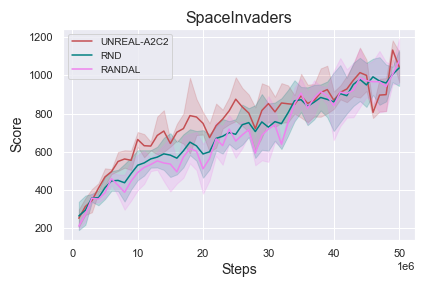}
	\includegraphics[width=0.32\textwidth]{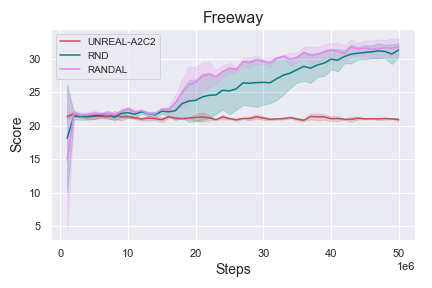}
	\includegraphics[width=0.32\textwidth]{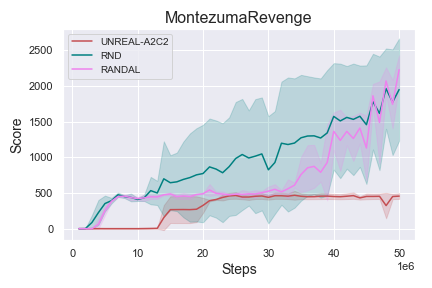}
	\caption[Atari experiments for RND vs UNREAL-A2C2 vs RANDAL agents]{Atari experiments for RND vs UNREAL-A2C2 vs RANDAL agents. All agents are superior in all environments over the traditional reinforcement learning agents and are equal in the dense reward task. However, the RND and RANDAL excel in the sparse reward tasks significantly outperforming the UNREAL-A2C2 agent in both the easy and hard search, sparse reward environments. RANDAL shows a less stable mean score than the RND, with large spikes in performance in successive validation scores, however shows a lower variance between the runs for both sparse reward tasks.}
	
	\label{Atari_RND}
\end{figure}

The UNREAL-A2C2 agent shows increased sample efficiency over the original implementation \cite[Figure~6]{Auxiliary} for MontezumaRevenge achieving the score of 500 at around 25 million steps as opposed to 50 million in \cite[Figure~6]{Auxiliary}. However, the UNREAL-A2C2 agent's performance plateaus at this score whilst the original implementation increases approximately linearly. The difference between these two agents could due to one of many changes. As using GAE is shown to generally increase performance \cite{GAE}, the suspect list is narrowed down to the inclusion of the previous action and reward to the policy as well as the use of a LSTM network. 

A clear parallel between these two methods can be seen when contrasting them from a \\ model-based/model-free viewpoint. Both RND and UNREAL agents can be said to utilise environment models to increase performance, as UNREAL agents learn a reward model $p(r_t|s_t)$, unconditioned on the action i.e.
\begin{equation}
p(r_t|s_t) = \sum_{a \in \mathcal{A}}p(a|s_t)(r_{t} | s_t, a) = \sum_{a \in \mathcal{A}}p(a|s_t)\mathcal{R}_s^a
\end{equation}  and RND agents learn a random feature model of the environment which can be said to be learning a simplistic density model of regions explored by the agent. This interplay between model-based and model-free reinforcement learning, seems to show increased sample efficiency and performance over purely model-free agents, again this is empirically supported by \cite{Imagine} as well as in Figure \ref{Atari_RND}. This theory is also backed by biological plausibility as model-based learning is said to be `ubiquitous' in animal brains \cite{ubiquity}. \\

Qualitative analysis of trained agents show that all sparse reward solutions explore the same 3 rooms in MontezumaRevenge, and the difference in the score is that the RND and RANDAL agents learn to use a collected sword to kill any enemy resulting in a large score increase. An interesting observation was that after the agents had reached the maximum total score, that they were capable of achieving, the RND agent aimlessly wanders the surrounding rooms and the UNREAL-A2C2 and RANDAL agents spend the remainder of the episode transitioning between two rooms. These can be explained as the RND agent is trying to maximise the intrinsic reward for the rest of the episode and the UNREAL-A2C2 agent being `interested' in the rooms transitions as they provide large pixel changes. The RANDAL agent shows a mix between these two behaviours, first moving back to the first room then for the remainder of the episode, it transitions between two rooms.

\section{Classic Control Experiments}
For the Classic Control Experiments the figures show the mean score $\pm$ the standard deviation for 5 different random starts. 
\subsection{Adam Hyperparameter Search}
Initial results from the Classic Control environments using hyperparameters from Table \ref{tbl:A2C Atari} and Table \ref{tbl:DDQN Atari} and Adam optimisation, showed relatively poor performance, so a small hyperparameter search was completed. For the A2C algorithm a random search was performed for 50 combinations of the entropy and learning rate with different random starts. The hyperparameters were sampled from a LogUniform($10^{-3},1$) and LogUniform($10^{-5},10^{-2}$) for the entropy and learning rate respectively.
\begin{figure}[H]
	\centering
	\includegraphics[width=.322\textwidth]{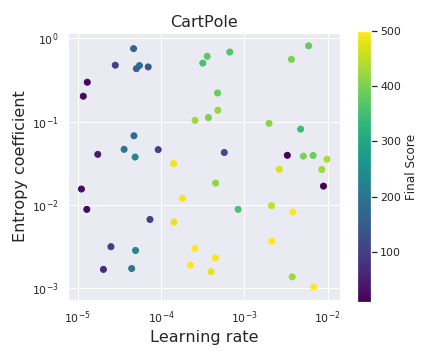}
	\includegraphics[width=.322\textwidth]{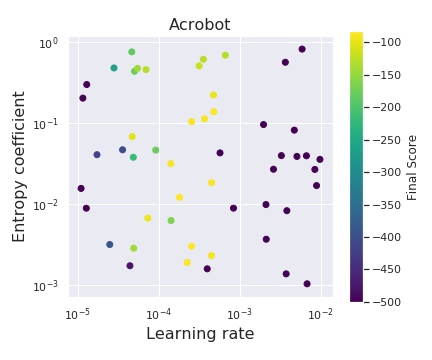}
	\includegraphics[width=.322\textwidth]{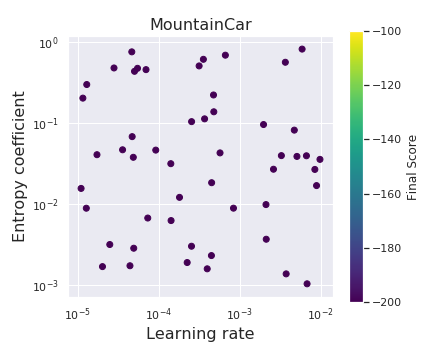}
	
	\caption[Classic Control A2C Adam hyperparameter search]{Classic Control A2C hyperparameter search, results show the Adam learning rate vs entropy with the colour representing the average of the last 5 validation scores of the agent, as this includes a measure of stability of the agent. It can be seen that learning rates that are optimal for both CartPole and Acrobot lie between the values of $2\times10^{-4} $ and $4\times10^{-4}$ and optimal entropy lies between $5\times10^{-3} $ and $2\times10^{-2}$. }
	\label{Control A2C hyper}
\end{figure}

The learning rates in Figure \ref{Control A2C hyper} have a larger affect on performance than the entropy coefficient, this is more pronounced in the Acrobot results where a large range of entropy values provide good performance. However even large entropy regularisation wasn't enough to for any the agents sampled to achieve any score in the sparse reward MountainCar task.\\

\noindent For the Classic Control Sync-DDQN agents, the exploration rate was linearly annealed over 80,000 steps, as to be proportional to the Atari experiments i.e. linearly annealed over ${1}/{25}th$ of training. These were done to try to keep the hyper-parameters consistent or proportional across all tasks as to reduce the number of variables that could be affecting the results. The agents are validated using a $\varepsilon_{test}$ of value 0.01 for all Sync-DDQN agents. The learning rate is sampled from a LogUniform($5\times10^{-5},10^{-2}$) distribution and the final $\varepsilon$ is chosen to be either 0.1 or 0.01 with equal probability. \\
\begin{figure}[H]
	\centering
	\includegraphics[width=.322\textwidth]{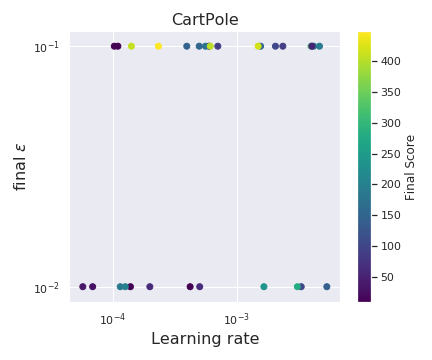}
	\includegraphics[width=.322\textwidth]{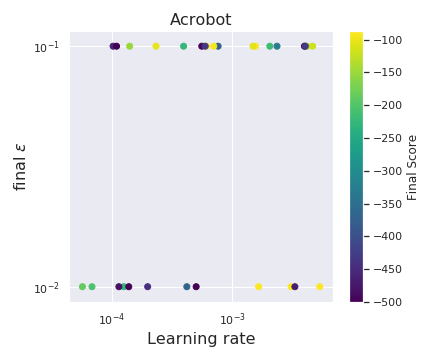}
	\includegraphics[width=.322\textwidth]{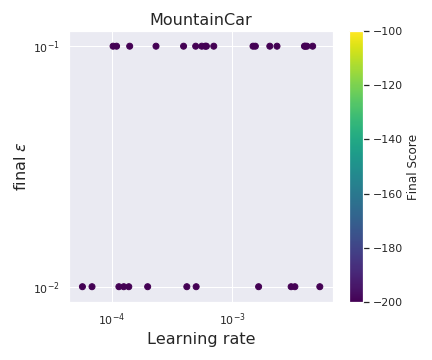}
	
	\caption[Classic Control Sync-DDQN Adam hyperparameter search]{Classic Control Sync-DDQN hyperparameter search, results show the Adam learning rate vs final $\varepsilon$ value over 30 samples, with the colour representing the average of the last 5 validation scores of the agent. The results are less clear than \ref{Control A2C hyper}, however it can be seen that both the CartPole and MountainCar agents suffer from poor exploration, as higher performance is seen for the higher final $\varepsilon$ value for CartPole and the MountainCar agents perform poorly, when they were expected to be able solve the task through sufficient random search.}
	\label{Control DDQN hyper}
\end{figure}

\subsection{Adam vs RMSProp}
For the A2C agent the learning rate and entropy coefficient were selected from the results shown in Figure \ref{Control A2C hyper}, to perform well over both the CartPole and Acrobot environments. These values were selected to be $2\times10^{-4}$ and $0.005$ for the learning rate and entropy coefficient respectively.

For the Sync-DDQN agent the only values shared that perform well in Figure \ref{Control DDQN hyper} for both CartPole and Acrobot. This is where the learning rate of $2\times10^{-4}$ and final exploration value $\varepsilon = 0.1$. As mentioned In Figure \ref{Control DDQN hyper}, the annealing the learning rate over 80,000 steps does not seem to provide enough exploration, so the number of steps were increased, so that the $\varepsilon$ was annealed from 1 to 0.1 over half of training i.e. 1 million steps. \\

\noindent The initial optimised Adam results, showed large dips in performance around the 500,000 step mark for both CartPole and Acrobot experiments, where the gradients have significantly overshot a local maximum, causing performance to drastically decrease. This was expected to be due to the momentum from the Adam optimiser causing the weights to be `slingshot' further from the global maxima. To test this, the RMSProp optimiser which has no momentum was used to compare the results, for the RMSProp agents the learning rates and entropy coefficient hyperparameters taken from the literature, as seen in Table \ref{tbl:A2C Atari} and Table \ref{tbl:DDQN Atari}, and the $\epsilon = 1\times 10^{-5}, \,\, \beta_v = 0.9$
\begin{figure}[H]
	\centering
	\begin{subfigure}[t]{\linewidth}
		\includegraphics[width=0.32\textwidth]{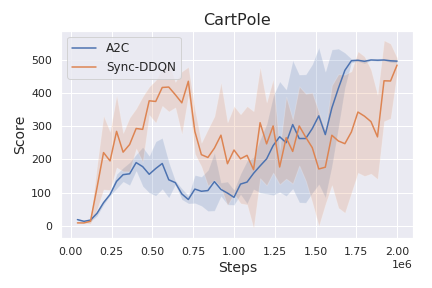}
		\includegraphics[width=0.32\textwidth]{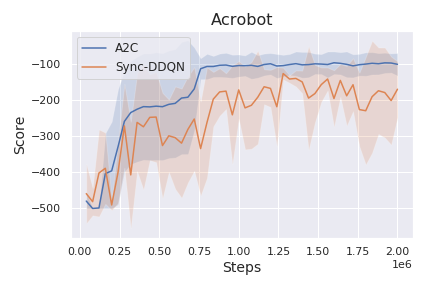}
		\includegraphics[width=0.32\textwidth]{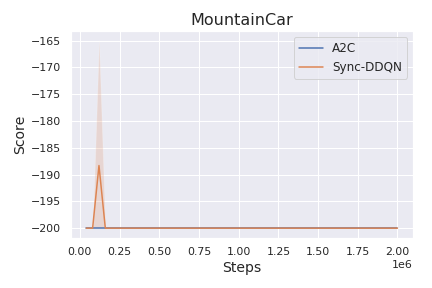}
		\subcaption{Optimised Adam Agents}
	\end{subfigure}
	\begin{subfigure}[t]{\linewidth}
		\includegraphics[width=0.32\textwidth]{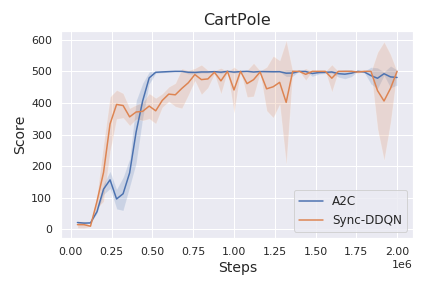}
		\includegraphics[width=0.32\textwidth]{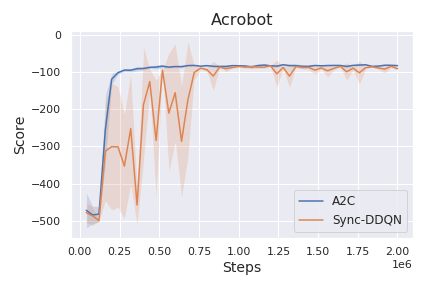}
		\includegraphics[width=0.32\textwidth]{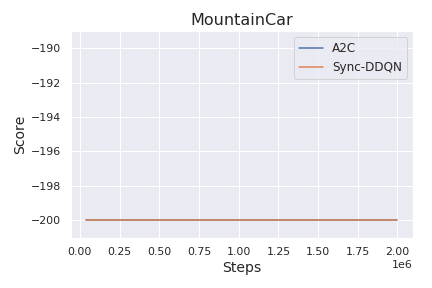}
		\subcaption{RMSProp Agents}
	\end{subfigure}
	\caption[Classic Control baselines Adam vs RMSProp for A2C and  Sync-DQQN agents]{Classic Control baselines Adam vs RMSProp for A2C and  Sync-DQQN agents. The results show that for low dimensional RL problems, RMSProp optimisation is significantly more stable and has faster convergence than Adam. The results also show that the policy gradient method is more stable than the value based method as the rewards fluctuate less over the course of training, matching up with the theory. However an unexpected result is the MountainCar environment which was expected to be solvable for the Sync DDQN through increased exploration over the initial results in Figure \ref{Control DDQN hyper}.}
	\label{Control_Baselines}
\end{figure}

\noindent Unlike the in Atari Freeway environment as seen in Figure \ref{Atari_Baselines}, both algorithms are unable to successfully navigate the MountainCar environment in Figure \ref{Control_Baselines}. Even with sufficient exploration for the Sync-DDQN agents the sparsity of the reward provides it very difficult for the agent to connect the series of actions to the reward received when reaching the target.  

\subsection{$\lambda$-Return and PPO}
For the Sync-DDQN($\lambda$) and A2C-GAE agents all hyperparameters were kept constant as increasing the step size decreased sample efficiency and performance, due to $n$=20 being much closer to the length of the episode ($200$ steps) in the Classic Control environments, compared to the Atari experiments. For simplicity and fairness, the PPO agents share many of the same hyperparameters of the A2C agents, and can be seen in Table \ref{Control_PPO}

\begin{table}[H]
	\centering
	\begin{tabular}{ c | c }
		Hyperparameter & Value \\
		\hline
		Optimiser & RMSProp \\
		Learning rate & $1\times 10^{-3}$  \\
		Number of actors & $32$  \\
		Entropy coefficient & $0.01$ \\
		Value coefficient & $0.5$ \\
		$n$-step period & $5$ \\ 
		Number of epochs & $4$ \\
		Number of minibatches & $1$ \\
		Discount factor $\gamma$ & $0.99$ \\
		PPO clip range & $\left[0.9,1.1\right]$ \\
		Gradient norm clipping & $0.5$ \\  
	\end{tabular}
	\caption[PPO Classic Control hyperparameters]{PPO Classic Control hyperparameters}
	\label{Control_PPO}
\end{table}

\begin{figure}[H]
	\centering
	\includegraphics[width=0.32\textwidth]{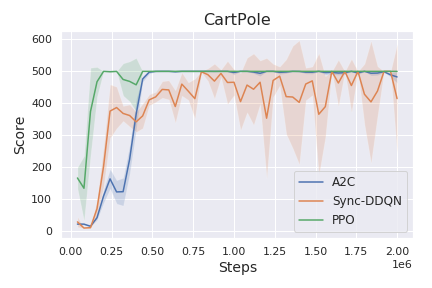}
	\includegraphics[width=0.32\textwidth]{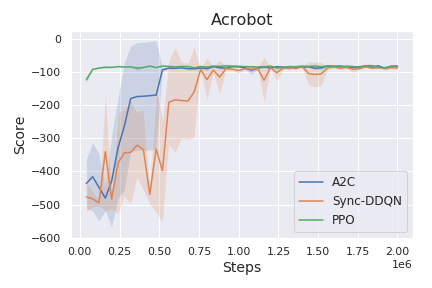}
	\includegraphics[width=0.32\textwidth]{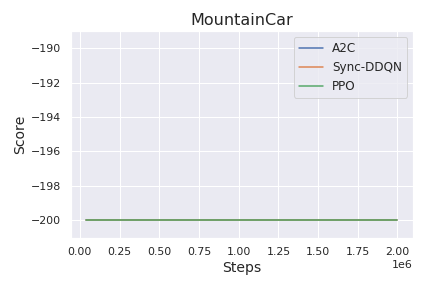}
	\caption[Classic Control A2C-GAE vs Sync-DDQN($\lambda$) vs PPO]{Classic Control A2C-GAE vs Sync-DDQN($\lambda$) vs PPO. The results show that the GAE slightly decreases sample efficiency for the A2C Acrobot agents, and Sync-DDQN CartPole agents. PPO, show greater sample efficiency and stability than all A2C and Sync-DDQN agents, achieving a perfect score from 0.5 million steps onwards for the CartPole environment.}
	\label{lambda_Control}
\end{figure}

The results from Figure \ref{lambda_Control} can be explained by the $R^\lambda$ and GAE, being more complex functions to model, thus slightly decreasing the learning speed of the agents.

\subsection{Sparse Reward Solutions}
Due to the nature of the pixel control task, the UNREAL-A2C and RANDAL agents only use the value replay and reward prediction tasks. All PPO-based agents (RANDAL, RND, ICM) use the same base policy hyperparameters as the Classic Control PPO agents seen in Table \ref{Control_PPO}, with the $\gamma_e$ and $\gamma_i$ and other algorithm specific hyperparameters from Tables \ref{tbl:RND Atari} and \ref{tbl:RANDAL Atari}.
\begin{figure}[H]
	\includegraphics[width=0.32\textwidth]{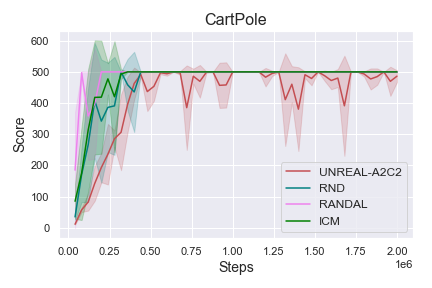}
	\includegraphics[width=0.32\textwidth]{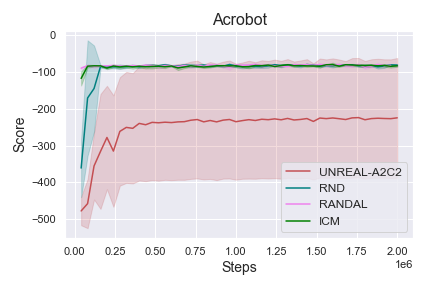}
	\includegraphics[width=0.32\textwidth]{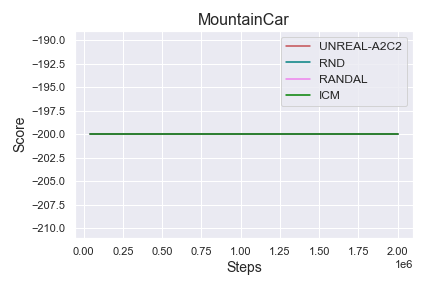}
	\caption[Classic Control experiments for RND vs UNREAL-A2C2 vs RANDAL agents]{Classic Control experiments for RND vs UNREAL-A2C2 vs RANDAL agents. The results surprisingly show no significant gain over the PPO agents, and the UNREAL-A2C2 Acrobot results show the importance of the affect that bad initialisation can have on deep RL agents.}
	\label{Control_sparse}
\end{figure}

Surprisingly Figure \ref{Control_sparse}, shows that auxiliary reward prediction and curiosity driven exploration are not enough for agents to sufficiently explore the MountainCar environment, this could be due to the normalisation techniques in curiosity agents or the intrinsic rewards are not enough to incentivise the agent to explore different methods of hill climbing before the parameters get stuck in a suboptimal stationary point. \\

\noindent The CartPole and Acrobot environments are clearly limiting benchmarks, easily solved by traditional agents, thus providing little information on the effectiveness on more complex algorithms. The MountainCar environment provides a difficult benchmark to reach for neural network parametrised agents. This is as neural networks often work better on higher dimensional data and require large amounts of data to converge compared to other models. The frequency of the reward is too small for these agents to sufficiently learn, another explanation is that the neural network agents are being stuck in a steep local stationary point where even after experiencing positive reward the update is not large enough to escape this point. One way to solve this problem would be to use a form of prioritised experience replay with large initial random exploration, to ensure the agent experiences enough rewards earlier on on the training.    

\chapter{Conclusions and Further Work} \label{chapter8}
Large performance gains in deep learning have often been attributed to increased computational power, however fundamental difficulties of sparse reward reinforcement learning have not been solved brute strength so far. Instead it has often relied on more intelligent exploration and exploitation algorithms, inspired by solutions found in nature, as reinforcement learning has had significant input from the fields of neuroscience and psychology. 

This report has shown that the combination and interplay of model-based and model-free reinforcement learning techniques in non-traditional ways, often increases sample efficiency and performance. This report has also shown that specifically finding reinforcement learning solutions that work well across multiple sparse reward tasks, generally increases performance across all tasks. 

\section{Further Work}
With the increased in performance of the UNREAL-A2C2, RND and novel RANDAL agents in sparse tasks, further work could be done to combine different aspects of these algorithms. For the UNREAL-A2C2 further investigation could be done to determine the compare the differences between this implementation vs the original, such as the use of GAE, Adam vs RMSprop optimisers, and the impact the feeding temporal information such as previous actions and rewards.

For the RANDAL agents a hyperparameter search could be done to optimise the weighting of the auxiliary tasks with the intrinsic reward. Combining intrinsic and extrinsic rewards for the reward prediction task is also an avenue for further exploration for the RANDAL architecture. \\

\noindent Following the successes of the interplay between model-based and model-free learning a proposed direction is to combine model-based imagination used in \cite{Imagine} with intrinsic motivation and auxiliary tasks.

\bibliographystyle{ieeetr}
\bibliography{mybibliography} 

\end{document}